\pdfoutput=1

\documentclass[11pt]{article}

\usepackage{emnlp2021}

\usepackage{times}
\usepackage{latexsym}

\usepackage[T1]{fontenc}

\usepackage[utf8]{inputenc}

\usepackage{microtype}

\usepackage{graphicx}
\usepackage{amsmath}
\usepackage{amsfonts}
\usepackage{amssymb}
\usepackage{booktabs}
\usepackage{multirow}
\usepackage{subfigure}
\usepackage[normalem]{ulem}
\usepackage[shortlabels]{enumitem}
\usepackage{multicol}
\usepackage{bigdelim}
\usepackage{rotating}
\usepackage{makecell}

\def\e{{\bf e}}

\def\p{{\bf p}}

\def\x{{\bf x}}

\def\0{{\bf 0}}
\def\1{{\bf 1}}

\def\CM{{\mathcal C}}
\def\DM{{\mathcal D}}

\def\LM{{\mathcal L}}
\def\MM{{\mathcal M}}

\def\OM{{\mathcal O}}

\newcommand{\nop}[1]{}

\usepackage{soul}
\usepackage{color}
\usepackage{xcolor}

\definecolor{purple-pink}{RGB}{255,172,233}
\definecolor{dark-green}{RGB}{174,224,146}
\definecolor{dark-blue}{RGB}{149,180,214}
\definecolor{yellow}{RGB}{254,253,175}
\definecolor{red-1}{RGB}{255,0,0}
\definecolor{green-1}{RGB}{0,176,80}
\newcommand{\hlc}[2][yellow]{{\sethlcolor{#1}\hl{#2}}}

\usepackage{color}

\title{Exophoric Pronoun Resolution in Dialogues with Topic Regularization}

\author{
Xintong Yu$^1$, Hongming Zhang$^2$, Yangqiu Song$^2$, Changshui Zhang$^1$, \\
\textbf{Kun Xu$^3$, and Dong Yu$^3$}\\
$^1$Institute for Artificial Intelligence, Tsinghua University (THUAI); \\
$^1$Department of Automation, Tsinghua University, Beijing, P.R.China \\
$^2$Department of CSE, The Hong Kong University of Science and Technology\\
$^3$Tecent AI lab\\
\texttt{yuxt16@mails.tsinghua.edu.cn, zcs@mail.tsinghua.edu.cn,} \\ 
\texttt{\{hzhangal, yqsong\}@cse.ust.hk,}
\texttt{\{kxkunxu, dyu\}@tencent.com}
}

\begin{document}
\maketitle
    \maketitle

    \begin{abstract}
        
        Resolving pronouns to their referents has long been studied as a fundamental natural language understanding problem.
        Previous works on pronoun coreference resolution (PCR) mostly focus on resolving pronouns to mentions in text while ignoring the exophoric scenario.
        Exophoric pronouns are common in daily communications, where speakers may directly use pronouns to refer to some objects present in the environment without introducing the objects first.
        Although such objects are not mentioned in the dialogue text, they can often be disambiguated by the general topics of the dialogue. 
        Motivated by this, we propose to jointly leverage the local context and global topics of dialogues to solve the out-of-text PCR problem.
        Extensive experiments demonstrate the effectiveness of adding topic regularization for resolving exophoric pronouns.

    \end{abstract}
    
    \section{Introduction}\label{sec:introduction}
     
    Grounding pronouns to objects they refer to is a challenging yet crucial natural language understanding problem.
    The coreference relationship between a pronoun and its referents is categorized into {\it endophora} and {\it exophora} based on whether the referred objects appear in text or out of text, and the former case can be further divided into {\it anaphora} if the referents appear in the preceding text of the pronoun and {\it cataphora} if in the following text~\cite{halliday1976cohesion,brown1983discourse}.
    Conventional studies on the pronoun coreference resolution (PCR) task in the NLP community  mainly focus on anaphora~\cite{hobbs1978resolving,nist2003ace,pradhan2012conll} and  some recent work analyzes cataphora in machine translation~\cite{DBLP:conf/acl/WongMH20},
    while mostly ignoring the exophoric pronouns.
    However, in daily dialogues or conversations, speakers may often use exophoric pronouns to refer to objects in the situational context that all speakers and listeners are aware of without introducing them in the first place. This limits the use of current PCR models in many real-world dialogue/conversation scenarios, e.g., text interpretation~\cite{hankamer1976deep,yule1979pragmatically} and downstream tasks such as dialogue generation~\cite{DBLP:conf/eccv/KotturMPBR18,DBLP:conf/cvpr/NiuZZZLW19}.

    \begin{figure}[t]
        \centering
        \includegraphics[width=\linewidth]{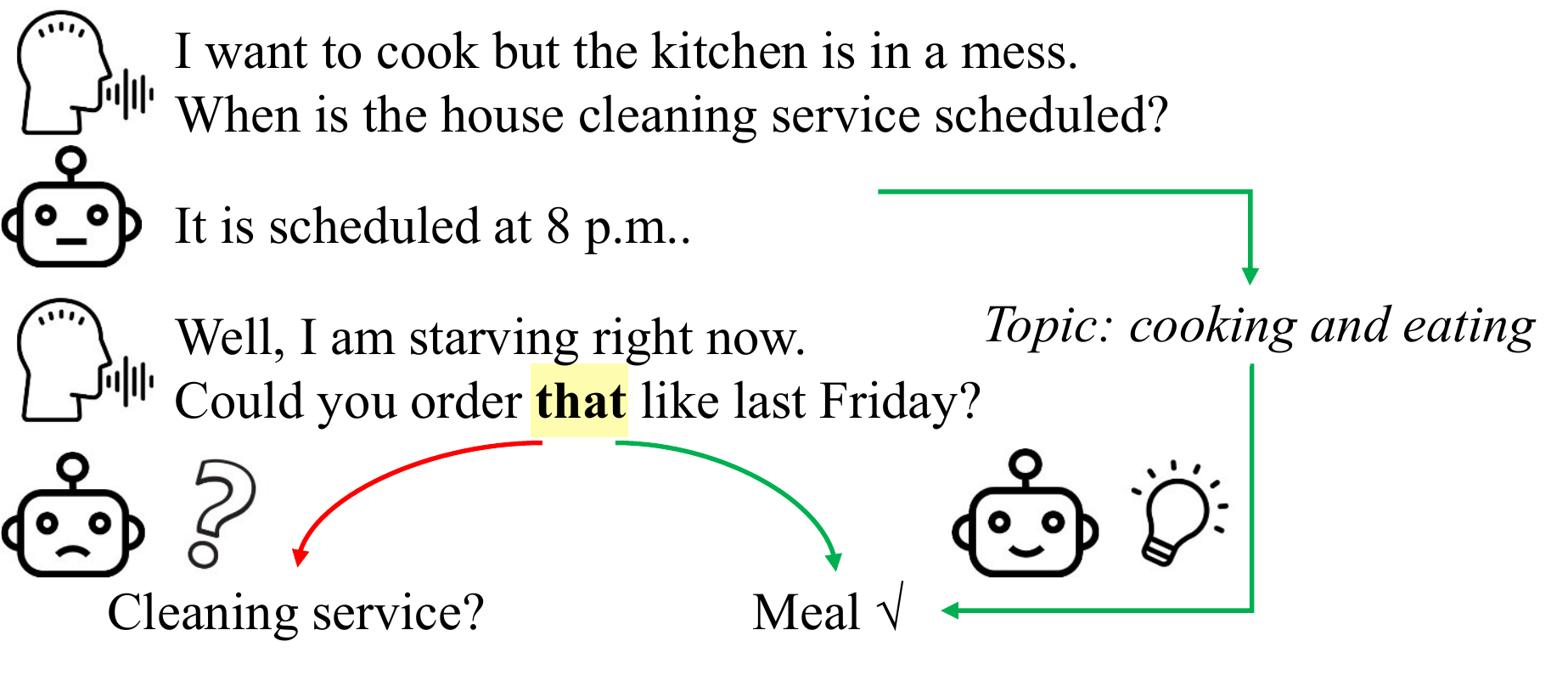}
        \caption{An example of resolving \hlc[yellow]{\textbf{exophoric pronouns}} in daily dialogues \textcolor{green-1}{with} and \textcolor{red-1}{without} the help of dialogue topics.}
        \label{fig:dialogue_example}
    \end{figure}

    Figure~\ref{fig:dialogue_example} shows an example of exophora.
    A person talks with his AI assistant (Siri/Alexa), ``Could you order that like last Friday?'' In this scenario, ``that'' is an exophoric pronoun whose referent can not be found in the dialogue text.
    A smart enough AI system should be able to resolve the pronoun ``that'' to some food rather than cleaning service based on the context.
    Such resolution of exophora is a crucial step in natural language understanding for the AI dialogue system to generate meaningful and relevant responses.
    
    Since traditional PCR tasks only focus on endophoric pronouns while ignoring exophoric ones, all existing models struggle when the correct referent is not in the textual context of the target pronoun. For example, most of the human-defined rules~\cite{hobbs1978resolving} (e.g., ``them'' can only refer to plural objects) and features~\cite{DBLP:conf/aaai/Ng05} (e.g., the distance between the target pronoun and candidate noun phrase) become either less effective or inapplicable in the exophoric setting.
    Unlike human-designed patterns or feature-based methods, the end-to-end coreference models~\cite{lee-etal-2018-higher,DBLP:conf/emnlp/JoshiLZW19} have the potential of resolving pronouns to external objects as long as the names of objects are provided as candidates.
    Nonetheless, these models heavily rely on the representation of local context produced by deep models so they always tend to resolve pronouns to the mentions in near text.
    As Figure~\ref{fig:dialogue_example} shows, the models could easily be distracted by the noun phrase ``cleaning service'' in text and resolve ``that'' to the service.

    To address the limitations of current models,
    we propose to take the overall dialogue topics into consideration.
    For the example in Figure~\ref{fig:dialogue_example}, 
    we can judge from the whole dialogue that the topic is about cooking and eating, so it is likely that the person needs some food. 
    If the AI system correctly resolves ``that'' to the topic-related out-of-text object ``meal,'' this may help the AI assistant to finally give a reasonable response, ``I will order the takeaway that you had last Friday.''

    To quantitatively define and evaluate exophora resolution, we leverage the VisPro dataset~\cite{DBLP:conf/emnlp/YuZSSZ19}, which annotates PCR information on visual dialogues. It is the only PCR dataset with annotations of out-of-text referents to the best of our knowledge. 
    While the original dataset provides images alongside dialogues, we observe that humans can resolve 96\% of exophoric pronouns in VisPro with only dialogue texts, which perfectly matches our research goal.
    Therefore, we perform out-of-text PCR experiments on the texts of VisPro.
    
    In this paper, we define the out-of-text PCR task and present a model, which jointly leverages the local context and global topics to better resolve pronouns to out-of-text objects.
    The model first identifies the overall dialogue topics and then assign larger scores to objects which are more relevant to the topics.
    By doing so, it less overfits the local context and learns to resolve pronouns based on global topics.
    Experimental results prove that the proposed model can significantly boost the performance of resolving exophoric pronouns without sacrificing the performance on in-text PCR.
    We also conduct an extensive analysis to show the contribution of different components.
    The data, code, and models are available at:
    \url{https://github.com/HKUST-KnowComp/Exo-PCR}.

    \section{Related Works}\label{sec:related-work}
    
    Coreference resolution is the task of identifying coreference relations among different mentions. As a vital natural language understanding component, a good coreference system could benefit many downstream tasks such as machine translation~\cite{DBLP:conf/eacl/Guillou12,DBLP:conf/acl/WongMH20}, 
    dialog systems~\cite{DBLP:conf/acl/StrubeM03}, 
    question answering~\cite{DBLP:conf/emnlp/DasigiLMSG19},
    and summarization~\cite{DBLP:journals/ipm/SteinbergerPKJ07}. 
    Due to the weak semantic meaning of pronouns~\cite{DBLP:conf/acl/Ehrlich81}, grounding pronouns to their referents (PCR) has been specially studied as a more challenging task than the general coreference resolution~\cite{mitkov1998robust,DBLP:conf/aaai/Ng05}.
    
    Previous PCR studies~\cite{DBLP:conf/aaai/Ng05,DBLP:conf/naacl/ZhangSS19} mostly focus on resolving pronouns to mentions in the near context. However, in informal text such as daily dialogues, it is common that pronouns may refer to out-of-text objects, which is crucial for dialogue understanding. 
    Such pronouns have long been discussed as ``pragmatically controlled anaphora'' in linguistics~\cite{hankamer1976deep,yule1979pragmatically,brown1983discourse}, but
    there has been few discussion of exophoric pronouns in the NLP community.
    \citet{DBLP:conf/emnlp/HangyoKK13} deal with exophora of zero pronouns, a special phenomenon in Japanese where an omitted argument of a predicate might refer to the “author” or the “reader” of the document.
    \citet{aktas-etal-2018-anaphora} qualitatively analyze exophoric reference in twitter conversations, where the antecedent of a pronoun could appear in the attached media or the quoted tweet.
    Unlike previous works, we follow a more general linguistics definition of exohpora~\cite{halliday1976cohesion} and evaluate it quantitatively.
    One recent work~\cite{DBLP:conf/emnlp/YuZSSZ19} annotates a dataset VisPro containing in-text and out-of-text referents for pronouns in Visual Dialog~\cite{DBLP:conf/cvpr/DasKGSYMPB17}, and solve the PCR task by involving visual information.
    In this work, we propose to resolve exophora in VisPro with texts as the only input. Our model jointly uses local context and global topic information for exophora resolution, which does not require the support of visual signals and thus can be applied to all scenarios.

    \section{The Task}\label{sec:task}
    
    In this section, we introduce details about the dataset construction and the task definition.

    \begin{figure}[t]
        \centering
        \includegraphics[width=\linewidth]{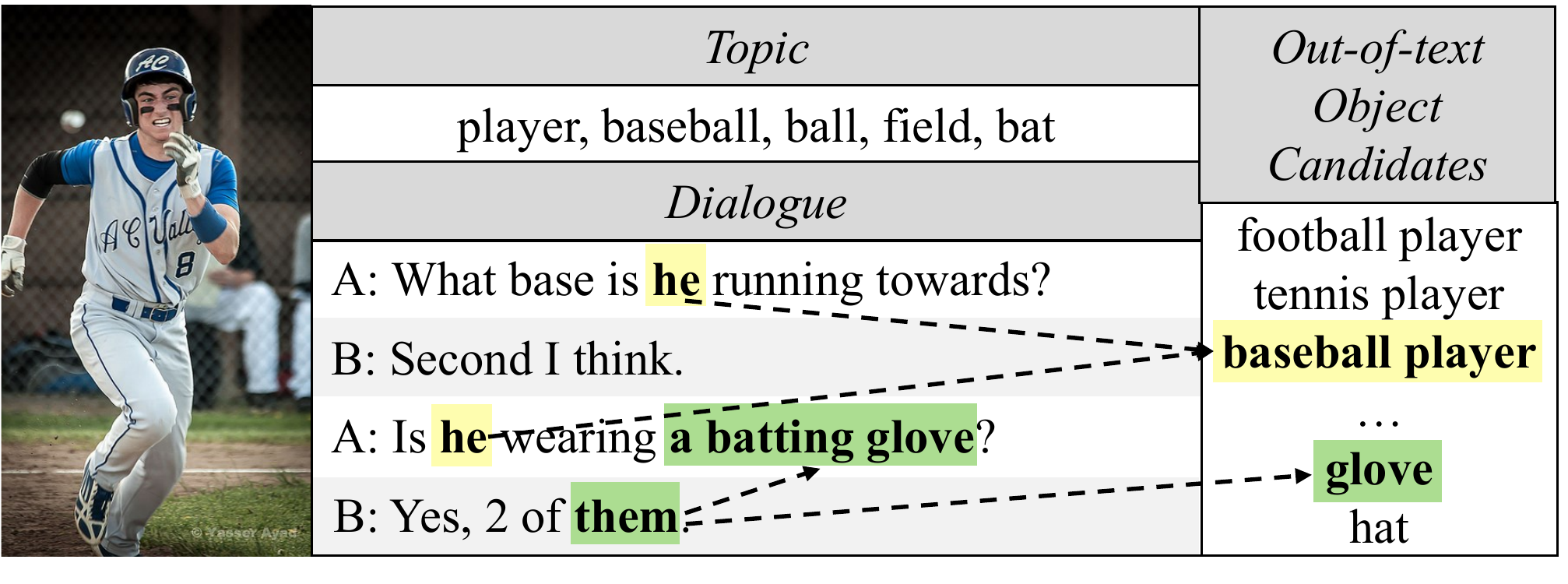}
        \caption{An example of the task.
        Pronouns are linked with their in-text and out-of-text referents.
        \hlc[yellow]{Exophoric} pronouns, \hlc[dark-green]{endophoric} pronouns, and their referents are marked with different colors.
        The topic words are predicted by an LDA model.
        }
        \label{fig:vispro_example}
    \end{figure} 
    
    \begin{figure*}[t]
        \centering
        \includegraphics[width=0.85\linewidth]{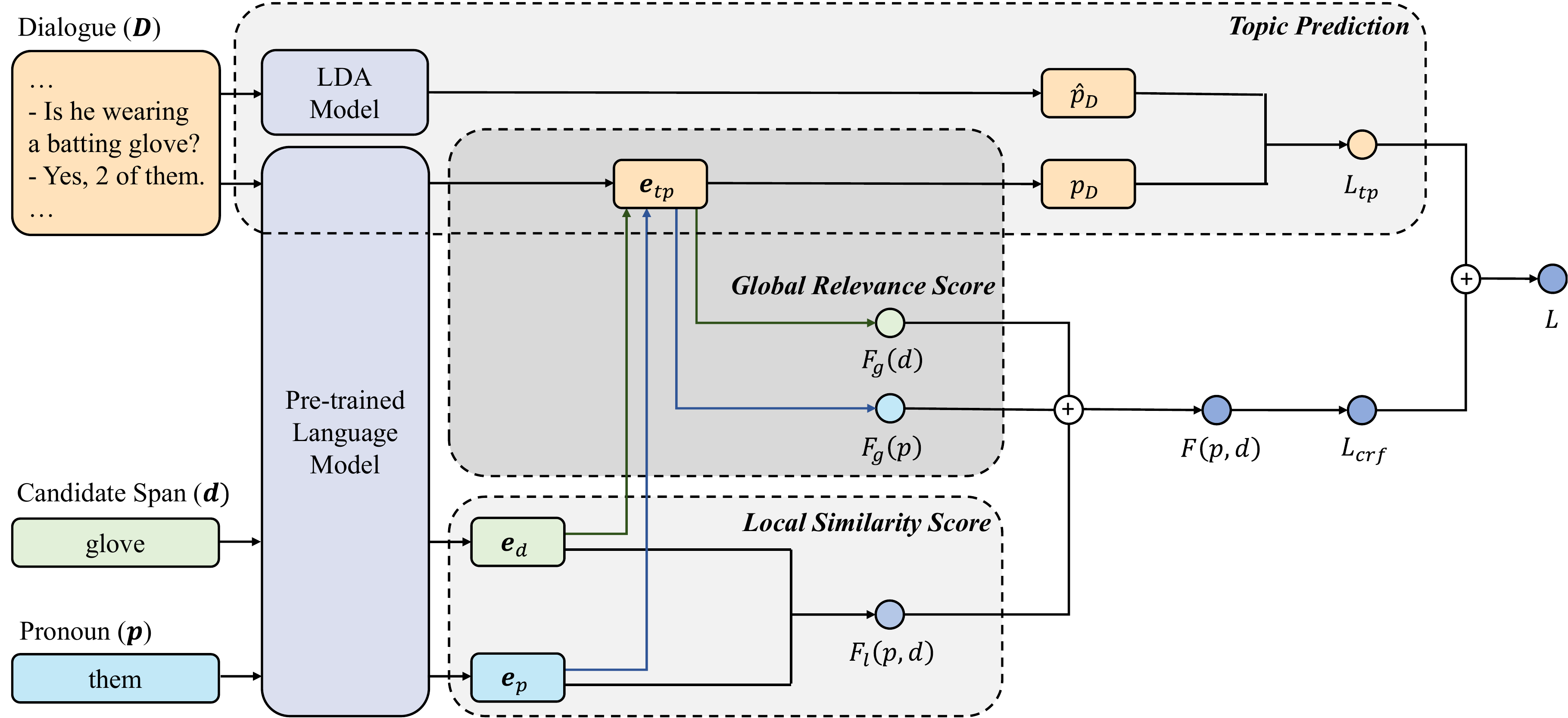}
        \caption{
        There are three main components in the proposed model: local similarity score calculation, global relevance score calculation, and topic prediction.
        The local score module calculates the similarity between a pronoun $p$ and a candidate span $d$ based on their textual representation.
        The global score module measures their relevance with the global dialogue topic.
        To help the topic embedding capture the topic information better, the topic prediction module uses the dialogue embedding to fit the topic vector predicted by LDA as an auxiliary task.
        }
        \label{fig:model}
    \end{figure*}
    
    \subsection{Dataset Setting}

    We construct the exophoric PCR dataset on top of VisPro~\cite{DBLP:conf/emnlp/YuZSSZ19}, which is the only dataset that provides rich exophoric pronoun annotations to the best of our knowledge. 
    Although the original research focus of VisPro is to study the importance of visual information in resolving pronouns in visual-related dialogues, we observe that in many cases, the dialogue text is enough for humans to make the correct resolution.
    Take Figure~\ref{fig:vispro_example} as an example. In the dialogue text, the pronoun ``he'' is exophoric because the referred person is not mentioned explicitly in the dialogue. 
    Even without the image, we can still guess that the dialogue is about a baseball game from clues like ``base'' and ``batting glove,'' and thus the pronoun ``he'' is more likely to refer to ``baseball player'' rather than other candidates.

    Quantitatively, we randomly select 100 exophoric pronouns in the development set of VisPro and find that 96\% of them can be correctly resolved without the visual information. Therefore, VisPro can be used as a valid dataset for the exophoric pronoun resolution task.
    A more detailed analysis is provided in Appendix~\ref{sec:appendix}.

    \subsection{Task definition}
    
    In this work, we focus on resolving pronouns to mentions inside dialogues and objects outside dialogues simultaneously.

    Given a pronoun $p$ in a dialogue $\DM$,
    we first select the noun phrases previous to $p$ in dialogue as candidates $\MM$ for in-text referents.
    For example, the noun phrases ``base'' and ``a batting glove'' are candidates of antecedents for ``them'' in Figure~\ref{fig:vispro_example}.

    To provide candidates for out-of-text referents for each dialogue, 
    \cite{DBLP:conf/emnlp/YuZSSZ19} randomly selects 30 noun phrases from image captions. However, such a setting is impractical when no caption is available (details are discussed in Appendix~\ref{sec:appendix}).
    As exophoric pronouns may refer to any object in daily life,
    we collect all the objects that frequently appear in the situational context of dialogues in VisPro to form an object pool $\OM$.
    The object pool contains 384 common object categories such as ``hat'' and ``glove'' shown in Figure~\ref{fig:vispro_example}. 
    The details of the collection are described in Sec~\ref{sec:dataset_processing}.

    The goal of the task is to identify the correct antecedents in $\MM$ and the correct out-of-text objects from $\OM$ by minimizing the loss:
    \begin{equation}\label{eq:objective1}
        \LM_{crf} = \LM_{i} + \LM_{o},
    \end{equation}
    where $\LM_{i}$ is the loss function for the in-text coreference resolution and $\LM_{o}$ for the out-of-text resolution. We then define them following the coreferenc loss in the end-to-end in-text coreference models~\cite{lee-etal-2018-higher}:
    \begin{equation}\label{eq:objective2}
    \begin{aligned}
    \LM_{i} &= - \log \frac{\sum_{c \in \CM_m}{e^{F(p, c;\DM)}}}{\sum_{m \in \MM}e^{F(p, m;\DM)}}, \\
    \LM_{o} &= - \log \frac{\sum_{c \in \CM_o}{e^{F(p, c;\DM)}}}{\sum_{o \in \OM}e^{F(p, o;\DM)}},
    \end{aligned}
    \end{equation}
    in which $F(\cdot)$ is the coreference score of pronouns $p$ with mentions $m$ or objects $o$,
    and $\CM_m$ and $\CM_o$ denote the correct referents in $\MM$ and $\OM$,  respectively.
    For instance, for the pronoun ``them'' in Figure~\ref{fig:vispro_example}, the model is required to not only recognize its antecedent in text to be ``a batting glove'' but also link it to ``glove'' in the external object pool.

    \section{The Model}\label{sec:model}

    The goal of the coreference model is to provide the coreference score $F(p, d)$ between a pronoun $p$ and a candidate $d$, which can either be a mention $m \in \MM$ or an external object $o \in \OM$.
    We divide the coreference score into three parts: the similarity score between $p$ and $d$ based on local context, the global topic relevance score of $p$,  and that of $d$:
    \begin{equation}
    F(p, d) = F_{l}(p, d) + F_{g}(p) + F_{g}(d).
    \end{equation}
    Specifically, $F_{l}$ calculates the similarity between $p$ and $d$ via local context representations, while $F_{g}$ acquires the relevance score between each text span and the global topics.

    To capture the topic information of the dialogues, we employ topic prediction as an auxiliary task of the model.
    The overall model architecture is shown in Figure~\ref{fig:model} and details are as follows.

    \subsection{Local Similarity Score}
    \label{subsec:pcr_in}
    Following \cite{DBLP:conf/emnlp/JoshiLZW19,lee-etal-2018-higher,DBLP:journals/corr/BahdanauCB14}, 
    for each span $s$, which could be either $p$, $m$, or $o$ and contains $T$ words $x_1, x_2, ..., x_T$, we first extract word embeddings from pre-trained language models as $\{\x_1, \x_2, ..., \x_T\}$.
    Then, we represent each span with the combination of the embeddings of the first token ($\x_1$), the last token ($\x_T$), the weighted sum of embeddings of all tokens in it ($\hat{\x}$), and the length feature of the span ($\phi(s)$):
    \begin{equation}
    \e_s = \left[\x_1, \x_T, \hat{\x}, \phi(s) \right],
    \end{equation}
    in which 
    \begin{equation}
    \begin{aligned}
    \hat{\x} &= \sum_{t=1}^{T} \alpha_t \cdot \x_t, \\
    \alpha_t &= \frac{\exp({\rm NN}_\alpha (\x_t)) }{\sum_{t=1}^{T} \exp({\rm NN}_\alpha (\x_t))}.
    \end{aligned}
    \end{equation}
    Here $[\cdot,\cdot]$ indicates the concatenation operation and ${\rm NN}$ the feed forward neural network.
    
    After acquiring the features of the spans, we then calculate the local similarity score between a pronoun $p$ and a candidate span $d$ as:
    \begin{equation}
    F_{l}(p, d) = {\rm NN}_r \left( \left[ \e_p, \e_d, \e_p \odot \e_d \right] \right).
    \end{equation}
    where $\odot$ denotes the element-wise multiplication.

    \subsection{Global Relevance Score}
    
    Although the out-of-text referents of exophoric pronouns are not mentioned in the text, they can be inferred from the dialogue context.
    As the subject of dialogue context, the dialogue topics play a vital part in exophora resolution.
    For the daily dialogue example in Figure~\ref{fig:dialogue_example}, we can infer from context words such as ``cook,'' ``kitchen,'' and ``starving'' that the dialogue topic is about cooking and eating, so the exophoric pronoun ``that'' is more likely to refer to ``meal'' rather than ``cleaning service.''
    
    Similarly, in the VisPro example in Figure~\ref{fig:vispro_example}, if we only read the sentence containing ``he,''
    it is hard to infer the targeting object of ``he'' to be a baseball player, a tennis player, or a football player.
    On the contrary,
    if we consider the whole dialogue as context, we can recognize the topic to be a baseball game, in which a man ``wearing a batting glove'' is ``running towards'' a ``base.'' Therefore, we can judge that this man must be a baseball player rather than a football or tennis player, so the exophoric pronoun ``he'' refers to the out-of-text object ``baseball player.''

    Based on the above observations, we leverage the overall dialogue topic to help grounding pronouns to out-of-text objects.
    To effectively encode the topic information of the whole dialogue, we first obtain the overall embedding $\e_{\DM}$ of a dialogue $\DM$ with pre-trained language models. 
    For LSTM-based models, we take the average embedding of all sentences as $\e_{\DM}$. For BERT-based models, we take the embedding of the special token [CLS]. Then we pass it through a feed forward neural network to obtain the dialogue topic embedding:
    \begin{equation}
    \e_{tp} = {\rm NN}_{tp} (\e_{\DM}).
    \end{equation}
    After that, to indicate the relevance between a span $s$ and the global topic of the dialogue, we calculate the topic relevance score as:
    \begin{equation}
    F_{g}(s) = {\rm NN}_{g} \left( \left[ \e_{tp}, \e_s, \e_{tp} \odot \e_s \right] \right).
    \end{equation}
    In the end, we calculate the final coreference scores of pronouns $p$ with in-text mentions $m$ and out-of-text objects $o$ as:
    \begin{equation}
    \begin{aligned}
    F(p, m) &= F_{l}(p, m) + F_{g}(p) + F_{g}(m), \\
    F(p, o) &= F_{l}(p, o) + F_{g}(p) + F_{g}(o).
    \end{aligned}
    \end{equation}
    
    With global relevance scores, models trained with VisPro are able to resolve exophoric pronouns based on dialogue topics. 
    In real-life scenarios such as Figure~\ref{fig:dialogue_example}, the key for understanding exophora is also the relevance between out-of-text objects and dialogue context. Thus the ability to resolve exophora with dialogue topics can also be transferred to such realistic cases.
    
    \subsection{Topic Prediction as Regularization}
    
    \begin{table}[t]
      \centering
      \scriptsize
        \begin{tabular}{l|l|l}
        \toprule
        No.   & LDA Topic Words & Summarized Topic \\
        \midrule
        15    & car, street, sign, road, vehicle & cars in streets \\
        16    & tree, grass, fence, animal, leaf & animals on grass \\
        23    & player, baseball, ball, field, bat & baseball game \\
        25    & kitchen, food, cut, stove, pot & kitchen \\
        28    & orange, banana, fruit, store, apple & fruit \\
        \bottomrule
        \end{tabular}%
      \caption{Example topics with five topic words extracted by the LDA model on the VisPro training set with $n_{tp}=40$. The last column presents topics summarized by human reading the extracted topic words.}
      \label{tab:LDA_topics}%
    \end{table}%
    
    To help the topic embedding $\e_{tp}$ better represent the topic information of the dialogue, we propose to use topic prediction as an auxiliary task.

    We first obtain the topic labels of dialogues by the most commonly used unsupervised topic model Latent Dirichlet Allocation (LDA)~\cite{DBLP:conf/nips/BleiNJ01}. 
    The LDA model extracts $n_{tp}$ topics from dialogues in the training set and represents each topic as a list of words with a high probability to appear under the topic. 
    Table~\ref{tab:LDA_topics} presents some topics of VisPro dialogues extracted by the LDA model. From the topic words, we can summarize that the No.15 topic is about cars in streets and that the No.25 topic discusses a kitchen.
    The topic label
    $\hat{\p}_{\DM}$ of a dialogue $\DM$ can be defined as:
    \begin{equation}
    \hat{\p}_{\DM} = {\rm LDA} ({\DM}) \in \mathbb{R}^{n_{tp}},
    \end{equation}
    where the $j^{\rm th}$ dimension of $\hat{\p}_{\DM}$ represents the probability of the dialogue corresponding to the No.$j$ topic.
    For instance, the LDA model predicts that the dialogue in Figure~\ref{fig:vispro_example} belongs to the No.23 topic in Table~\ref{tab:LDA_topics} with 60\% possibility and thus the $23^{\rm th}$ dimension of $\hat{\p}_{\DM}$ is 0.6.
    
    As $\hat{\p}_{\DM}$ sums up to 1 and each dialogue could associate with several topics, we fit $\hat{\p}_{\DM}$ by $\e_{tp}$ with a L2 loss after a softmax function\footnote{We also tried other loss functions, such as KL-divergence between $\p_{\DM}$ and $\hat{\p}_{\DM}$, and cross entropy loss after a sigmoid function for each dimension of ${\rm NN}_p (\e_{tp})$. Empirical studies show that the L2 loss achieves the best performance.}:
    \begin{equation}
    \begin{aligned}
    \p_{\DM} &= {\rm softmax}\left( {\rm NN}_p (\e_{tp}) \right), \\
    \LM_{tp} &= \frac{1}{2} ||\p_{\DM} - \hat{\p}_{\DM}||_2^2.
    \end{aligned}
    \end{equation}
    We use the topic prediction loss as a regularization term to the total loss:
    \begin{equation}
    \begin{aligned}
    \LM = \LM_{crf} +  \LM_{tp} = \LM_{i} + \LM_{o} + \LM_{tp},
    \end{aligned}
    \end{equation}
    where $\LM_{i}$ and $\LM_{o}$ are defined in (\ref{eq:objective2}).
    As a result, the final loss function $\LM$ can be optimized in an end-to-end manner.
    
    \section{The Experiment}\label{sec:experiment}
    In this section, we introduce the experiment details.
    
    \subsection{Dataset}\label{sec:dataset}

    We use VisPro~\cite{DBLP:conf/emnlp/YuZSSZ19} as the dataset, which contains 4,000 train, 500 development, and 500 test dialogues.
    The train, development, and test sets of VisPro contain 13,686, 1,726, and 1,781 pronouns with out-of-text referents and 13,986, 1,742, and 1,756 pronouns with in-text antecedents, respectively.
    
    \subsection{Evaluation Metrics}

    We use different metrics for in-text and out-of-text PCR due to the different numbers of candidates.
    For the in-text PCR, each pronoun has 10.3 candidates and 1.6 correct referents on average. Thus we follow the previous work~\cite{DBLP:conf/emnlp/YuZSSZ19} to employ Precision (P), Recall (R), and F1 score as the evaluation metrics.
    For the out-of-text PCR, as all 384 common object nouns are candidates and only one of them is correct, the F1 score is no longer suitable.
    For example, if the model predicts the correct answer to be the second place out of 384 candidates, it means that model can somehow understand the pronoun, while the F1 metric will count it as wrong.
    Therefore, we view out-of-text PCR as a ranking problem, where objects that a pronoun refers to should have a higher rank, and evaluate all models by the recall at 1, 5, and 10.

    \subsection{Baselines}

    We add our global relevance score module and topic prediction module on basis of the following end-to-end coreference resolution models which only contains the local similarity score module\footnote{
    We do not compare with CorefQA~\cite{DBLP:conf/acl/WuWYWL20} because it selects in-text antecedents as a reading comprehension task, which cannot be applied to out-of-text objects. We do not compare with VisCoref~\cite{DBLP:conf/emnlp/YuZSSZ19} because it requires images as input, while our setting is text-only.
    }:
    
    \begin{itemize}[leftmargin=*]
        \item End-to-end model with \textbf{LSTM} based on \textbf{ELMo} embedding~\cite{lee-etal-2018-higher}
        , which extracts features by a BiLSTM upon ELMo embeddings.
        \item End-to-end model with \textbf{BERT} embedding~\cite{DBLP:conf/emnlp/JoshiLZW19}.
        \item End-to-end model based on \textbf{SpanBERT} embedding~\cite{DBLP:journals/tacl/JoshiCLWZL20}
       , which can better represent text spans.
    \end{itemize}

    \begin{figure}[t]
        \centering
        \includegraphics[width=\linewidth]{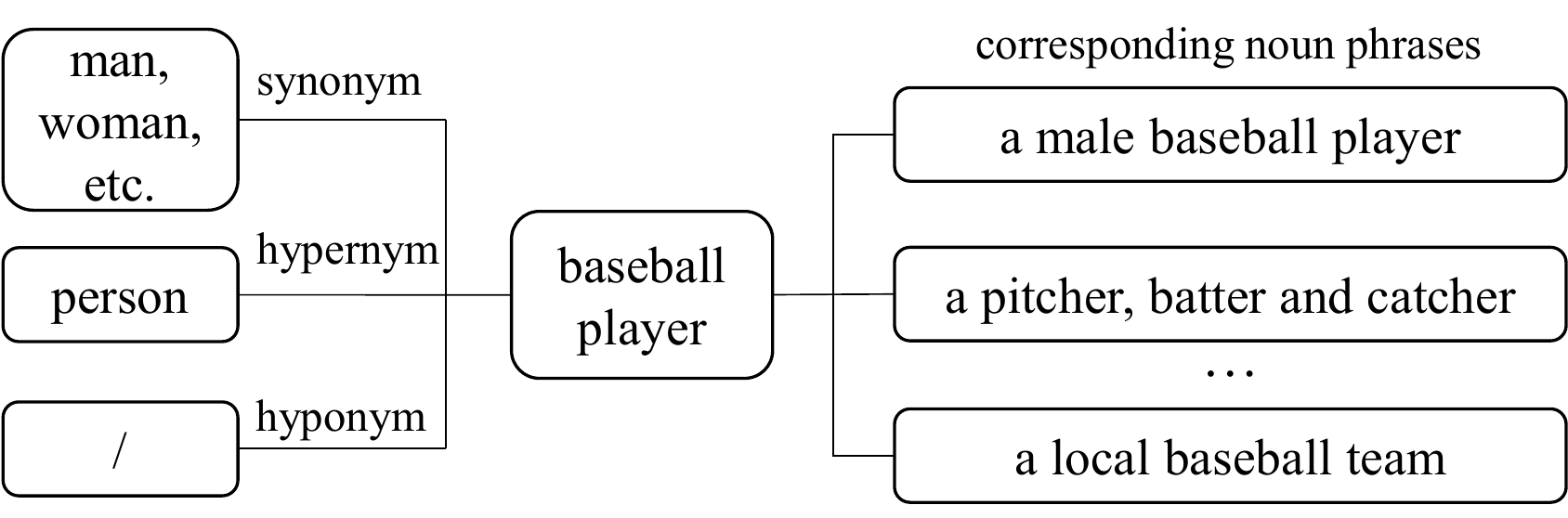}
        \caption{An example of the object category ``baseball player.'' Each object category contains its synonyms, hypernyms, hyponyms, and corresponding noun phrases.}
        \label{fig:object}
    \end{figure} 

    \begin{table*}[t]
    \small
        \centering
        \begin{tabular}{l|ccc|ccc|ccc}
        \toprule
        \multicolumn{1}{c|}{\multirow{3}[3]{*}{Model}} & \multicolumn{6}{c|}{Out-of-text PCR}       & \multicolumn{3}{c}{In-text PCR} \\
    \cmidrule{2-10}          & \multicolumn{3}{c|}{Not Discussed} & \multicolumn{3}{c|}{Discussed} &       &       &  \\
              & R@1   & R@5   & R@10  & R@1   & R@5   & R@10  & P     & R     & F1 \\
        \midrule    
        ELMo-LSTM & 61.54 & 66.19 & 66.80 & 70.86 & 71.25 & 71.25 & 88.15 & 66.05 & 75.51 \\
        \@  + topic (ours) & 68.02 & 71.66 & 72.06 & 72.49 & 72.96 & 72.96 & 87.55 & 70.43 & 78.06 \\
        \midrule
        BERT-base & 87.45 & 89.68 & 90.49 & 89.74 & 94.64 & 94.79 & 86.51 & 80.63 & 83.47 \\
        \@  + topic (ours) & \textbf{90.49} & \textbf{93.72} & \textbf{95.75} & 92.46 & 96.43 & 96.89 & 85.79 & 83.66 & 84.72 \\
        \midrule    
        SpanBERT-base & 87.65 & 92.11 & 92.51 & 91.38 & 94.25 & 94.79 & \textbf{89.08} & 79.35 & 83.94 \\
        \@  + topic (ours) & 90.28 & 93.32 & 93.93 & \textbf{93.63} & \textbf{96.50} & \textbf{97.05} & 83.97 & \textbf{85.78} & \textbf{84.87} \\
        \bottomrule
        \end{tabular}%
        \caption{Results of experiments for out-of-text PCR evaluated by Recall (R) in the top 1, 5, and 10 predictions and in-text PCR measured by Precision (P), Recall (R), and F1 score. The best results are shown in \textbf{bold} font.}  
        \label{tab:result}%
    \end{table*}%

    \subsection{Implementation}\label{sec:implementation}
    \label{sec:dataset_processing}
    
    \noindent \textbf{Dataset Processing:}
    To collect common object categories in VisPro, we first map 2,600 noun phrases annotated as out-of-text referents in VisPro to a compact list of 384 object categories by removing all modifiers and merging similar phrases.
    For instance, pronouns referring to ``a male baseball player'' or ``a local baseball team'' are both mapped to the object ``baseball player.''
    Moreover, some objects have similar or overlapping meanings with other objects (e.g., ``pond'' similar to ``pool'') but only one is labeled as the gold answer of a pronoun. It would be problematic if we directly label all others as wrong.
    To solve this problem, we use the synonyms, hypernyms, and hyponyms obtained from synset in Wordnet~\cite{DBLP:journals/cacm/Miller95} in NLTK~\cite{DBLP:conf/acl/Bird06} as extra information attached to each object category. If a pronoun refers to a particular object in the external object pool, then the synonyms, hypernyms, and hyponyms of the targeting object are masked during the training and testing process. An example of an object category ``baseball player'' is shown in Figure~\ref{fig:object}. 
    Note that other person categories which are not a synonym, hypernym or hyponym of ``baseball player'', such as ``tennis player'' and ``football player'', are not masked.
    
    Last but not least, we split the pronouns with out-of-text referents by whether a pronoun simultaneously refers to some mentions in the dialogue.
    If a pronoun has both in-text and out-of-text referents, such as ``them'' in Figure~\ref{fig:dialogue_example}, which refers to ``a batting glove'' in the dialogue as well as ``glove'' in the object pool, we denote it as ``Discussed'' in the dialogue.
    If a pronoun only has out-of-text referents, such as ``he'' in Figure~\ref{fig:dialogue_example}, which only refers to the object ``baseball player,'' we denote it as ``Not Discussed'' in the dialogue.
    While ``Not Discussed'' pronouns strictly match the definition of exophora, grounding the ``Discussed'' pronoun to out-of-text objects is also an important step towards linking dialogue text to the environment.
    In VisPro, 25.02\% of all pronouns with out-of-text referents are ``not discussed.''

    \noindent \textbf{Training Details:}
    We follow the hyperparameters set in \cite{DBLP:conf/emnlp/JoshiLZW19}. 
    The number of topics $n_{tp}$ is set to 40 for LDA.
    The topic prediction module in the model contains one hidden layer of size 1,000.
    Gold mentions are provided for training and testing of the models.
    During testing, the in-text antecedents are chosen in the same way as \cite{lee-etal-2018-higher}. 
    For the out-of-text part, objects $o$ with scores $F(p, o) > 0$ are deemed as the prediction of out-of-text referents for the pronoun $p$ and the selected objects are ranked according to the scores.
    Models are trained for ten epochs, and the best ones are selected based on their performance on the development set.

    \section{The Results}\label{sec:result}

    

    From the experimental results in Table~\ref{tab:result}, we can observe that 
    BERT and SpanBERT based models outperform ELMo-LSTM based models, which is consistent with the observation in~\cite{DBLP:conf/emnlp/JoshiLZW19} mainly because of their stronger context representation ability.
    On top of them, incorporating global topics improves recall for both exophoric and endophoric pronouns.
    Last but not least, for in-text PCR, adding topic information only slightly influences the precision while significantly improving the recall.
    As a result, it also achieves better overall F1 performance.

    Further analyzing the performances of models on out-of-text PCR, we observe that the ``Not Discussed'' pronouns are more challenging than the ``Discussed'' group for all models. This makes sense because if a pronoun refers to some noun phrases in text, the embedding of the pronoun will encode the information of those noun phrases via the language models. 
    For instance, if the representation of ``them'' in Figure~\ref{fig:vispro_example} encodes the context ``a batting glove,'' it would be easier to identify the semantically related object ``glove'' as the out-of-text referent.
    In contrast, ``Not Discussed'' pronouns do not have any noun phrase antecedent in the dialogue and are thus more challenging.
    In such cases, the effect of incorporating global semantics becomes more significant than in ``Discussed'' cases.
    In the rest of this section, we present a detailed analysis with the BERT-base + topic model, which achieves the highest performance on ``Not Discussed'' pronouns and comparable performances on other settings, to show when our model performs well and when it fails.

    \begin{table}[t]
        \centering
        \small
        \begin{tabular}{l|c|ccc}
            \toprule
            \multicolumn{1}{c|}{Model} & Object Type & R@1   & R@5   & R@10 \\
            \midrule
            \multirow{2}{*}{BERT-base} & Infrequent & 39.66 & 51.72 & 53.45 \\
            & Frequent & 93.81 & 94.72 & 95.41 \\
            \midrule
            \multirow{2}{*}{\@ + topic} & Infrequent & 46.55 & 65.52 & 74.14 \\
            & Frequent & 96.33 & 97.48 & 98.62 \\
            \bottomrule
        \end{tabular}%
        \caption{Recall of ``Not Discussed'' pronouns in the test set referring to ``Infrequent'' and ``Frequent'' objects.}
        \label{tab:freq}%
    \end{table}%

    \begin{figure}[t]
        \centering
        \includegraphics[width=0.4\textwidth]{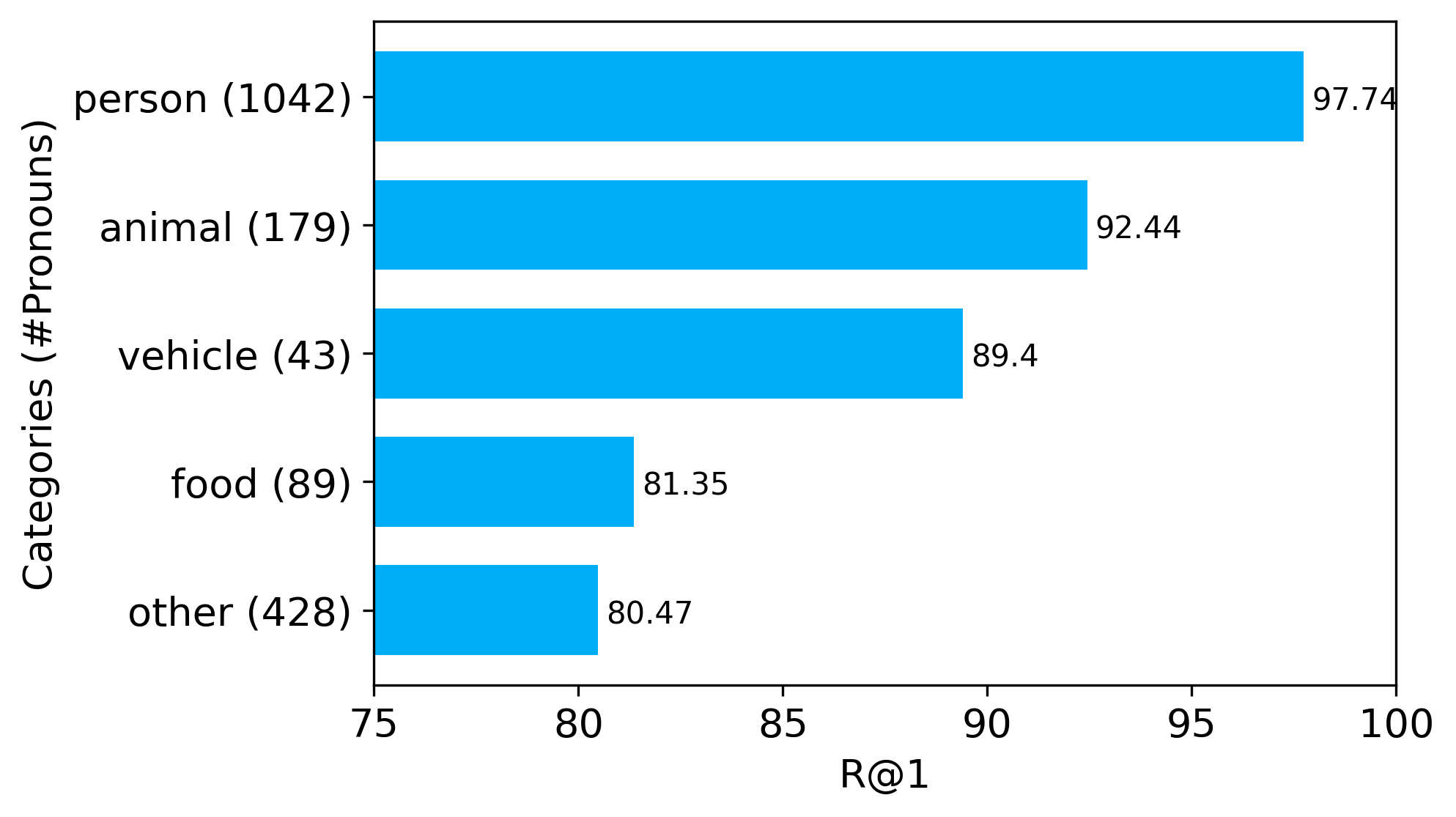}
        \caption{Performance and number of pronouns in the test set related to different out-of-text object categories.}
        \label{fig:cat}
    \end{figure}

    \subsection{Influence of Frequency}
    \label{sec:frequency}

    In the external object pool, the appearances of different objects varies. For instance, ``man'' appears 3,084 times in the training set, while ``monkey'' only appears once. 
    To investigate the influence of such imbalance, we split the object list by their occurrence frequency, with occurrence less than 50 times as ``Infrequent'' objects, which make up 85.1\% of list, and the rest as ``Frequent'' objects.

    As observed in Table~\ref{tab:freq}, performances on infrequent objects are much lower than frequent ones, which indicates that although the models achieve high scores on frequent objects, they still fail to do well on the majority of relatively rare objects.
    This observation also shows that the exophoric PCR problem is still far from being solved.
    Compared to models focusing on local information, the proposed model, which incorporates the overall topics, boosts the performance by a large margin, especially on infrequent pronouns.
    
    \subsection{Influence of Object Categories}

    Besides the influence of frequency, we are also interested in how well our model can perform on different object categories. 
    We record the performance of pronouns related to the four most common categories\footnote{Here a pronoun is deemed as related to a major category if the object it refers to is exactly that category or a hyponym of the category. For example, pronouns linked to ``person'' or ``man'' are both considered related to ``person.'' We also report the number of related pronouns in the test set.} (person, animal, vehicle, and food) in Figure~\ref{fig:cat}, from which we can see that pronouns related to ``person'' and ``animal'' are most common and easiest to be resolved, which is consistent with our previous observation that our model performs better on frequent objects than on infrequent ones.

    \begin{table}[t]
        \centering
        \small
        \begin{tabular}{l|cc|cc}
            \toprule
            \multicolumn{1}{c|}{}  & \multicolumn{2}{c|}{Out-of-text} & \multicolumn{2}{c}{In-text} \\
            & R@1   & $\Delta$R@1 & F1    & $\Delta$F1 \\
            \midrule
            Our full model & 90.49 &     -  & 84.72 & - \\
            \midrule
            \@ - topic prediction & 88.46 & -2.02 & 84.08 & -0.63 \\
            \@ - masking synonyms & 56.88 & -33.60  & 84.04 & -0.67 \\
            \@ - in-text training & 87.25 & -3.24 & 25.12  & -59.60  \\
            \@ - out-of-text training & 48.99 & -41.50 & 82.47 & -2.24 \\
            \bottomrule
        \end{tabular}%
        \caption{Ablation study results.}
        \label{tab:ablation}%
    \end{table}%
    
    \subsection{Ablation Study}

    We present the ablation study in Table~\ref{tab:ablation}, from which we can see that all components contribute to the ultimate success.
    For example, performance drops when removing the topic prediction loss as regularization, which indicates that the topic prediction module can help the embedding of the dialogue to capture the topic information better.
    Besides that, if we do not mask out the synonyms, hypernyms, and hyponyms of the object categories during training, 
    the performance drops dramatically.
    It shows the importance of masking possible distractions to provide unique labels during training.
    Last but not least, one contribution of the proposed model is the joint training of both the in-text and out-of-text PCR and, the results show that removing either of them in the training process will result in a performance drop on both tasks.
    Similar improvement by joint training is also observed in \cite{DBLP:conf/eacl/BaiZSX21}, where the in-text PCR task is jointly trained with the character linking task that links the endophoric pronouns in TV show scripts to the characters.

    \subsection{BERT-base VS BERT-large}
    
    \begin{table}[t]
        \centering
        \small
        \begin{tabular}{l|cc|c}
        \toprule
        \multicolumn{1}{c|}{\multirow{3}[4]{*}{Model}} & \multicolumn{2}{c|}{Out-of-text} & \multicolumn{1}{c}{\multirow{2}[2]{*}{In-text}} \\
              & \multicolumn{1}{l}{Not Discussed} & Discussed &  \\
    \cmidrule{2-4}          & R@1   & R@1   & F1 \\
        \midrule
        BERT-base & 87.45 & 89.74 & 83.47 \\
        \@  + topic & \textbf{90.49} & 92.46 & 84.72 \\
        \midrule
        BERT-large & 87.25 & 90.83 & 84.62 \\
        \@  + topic & 88.46 & 92.00 & 85.08 \\
        \midrule
        SpanBERT-base & 87.65 & 91.38 & 83.94 \\
        \@  + topic & 90.28 & \textbf{93.63} & 84.87 \\
        \midrule
        SpanBERT-large & 87.65 & 91.61 & 86.64 \\
        \@  + topic & 89.68 & 93.40 & \textbf{87.22} \\
        \bottomrule
        \end{tabular}%
        \caption{Performance comparison among BERT-base, BERT-large, SpanBERT-base, and SpanBERT-large embeddings. The best results are in \textbf{bold} font.}    
        \label{tab:bert-large}%
    \end{table}%
    
    Table~\ref{tab:bert-large} compares the performance of models based on BERT-base, BERT-large, SpanBERT-base, and SpanBERT-large. Incorporating topic information consistently improves performance on out-of-text PCR for all models while achieving comparable scores on the in-text one. Besides, we surprisingly find out that compared to BERT-base and SpanBERT-base, even though BERT-large and SpanBERT-large achieve higher scores on in-text PCR, their performance on the out-of-text PCR slightly drops. An explanation is that they may easily overfit the local context and ignore the global topic information due to their deep model.

    \subsection{Case Study}
    
    \begin{figure}[t]
        \centering
        \includegraphics[width=0.47\textwidth]{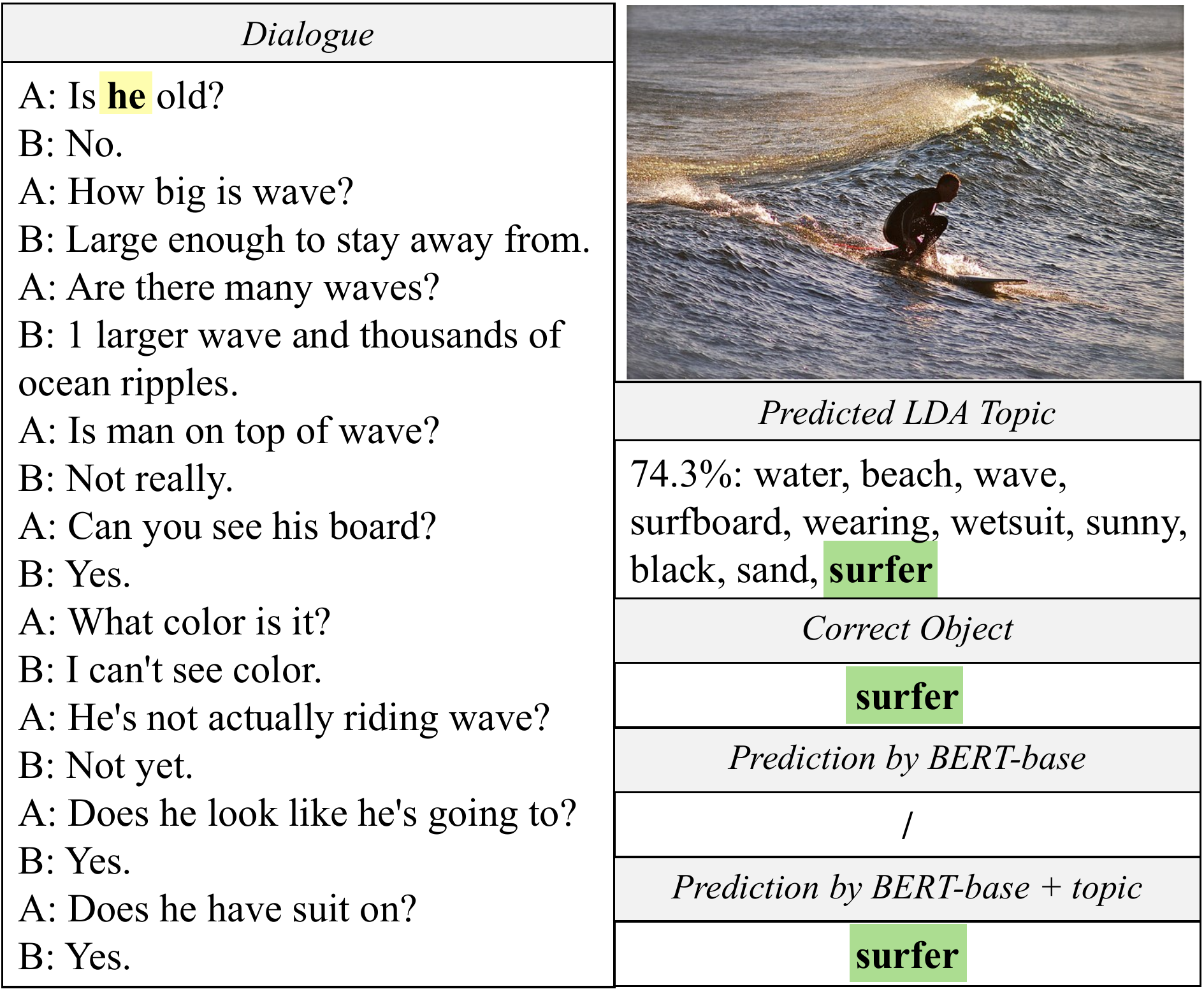}
        \caption{Case study for out-of-text PCR. 
        \hlc[yellow]{Target pronouns} and \hlc[dark-green]{correct out-of-text objects} with their hints are marked in different colors. 
        Note that we only show the corresponding images here for clarity and that they are \textbf{not} provided to the models.
        }
        \label{fig:case}
    \end{figure}

    Figure~\ref{fig:case} shows a dialogue about a male surfer. The referents of the pronoun ``he'' is ``Not Discussed'' in the dialogue text.
    The model that can only access the local context cannot identify any object related to the pronoun. In contrast, the model with topic prediction assigns a high probability of 74.3\% for the topic of the dialogue to be surfing judging from the related words such as ``wave'' and ``board.'' Thus it identifies ``surfer'' as the out-of-text referent for the pronoun.
    More cases are shown in Appendix~\ref{sec:appendix_case}.

    \subsection{Error Analysis}
    
    \begin{figure}[t]
        \centering
        \includegraphics[width=0.37\textwidth]{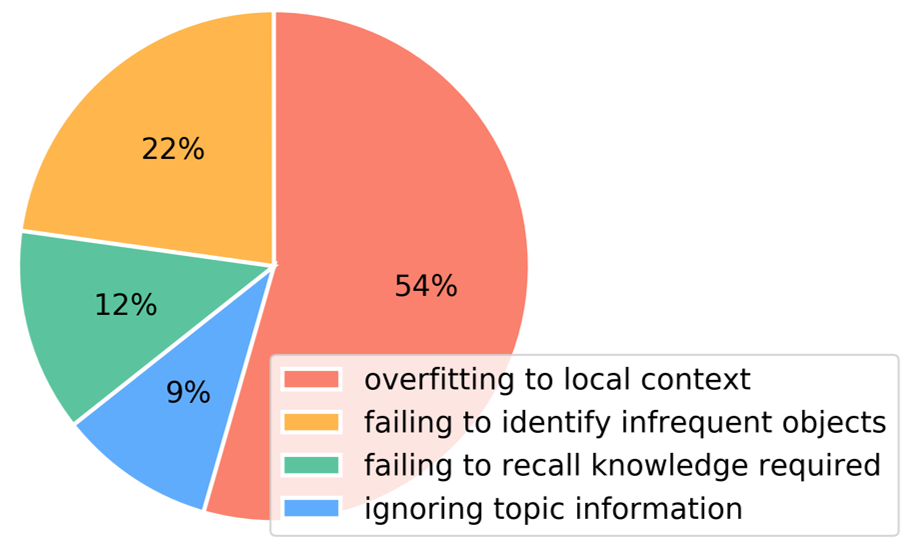}
        \caption{Error distribution in out-of-text PCR.}
        \label{fig:error_types}%
    \end{figure}

    We first quantitatively study the error types of the BERT-base + topic model by randomly selecting 60 mistaken predictions in out-of-text PCR, including 30 cases for the ``Not Discussed''  pronouns and 30 for the ``Discussed'' ones.
    We observe that 1/3 of the cases are also difficult for humans to identify the correct objects without access to the corresponding images. This is either because the dialogue text does not contain enough clues to infer the right answer, or multiple answers are reasonable but only one is annotated.
    For the other 2/3 cases, Figure~\ref{fig:error_types} shows that more than half of errors are still from overfitting to local context and 10\% from failure to use the topic information. Other 23\% errors come from failure to associate pronouns with infrequent objects as discussed in Section~\ref{sec:frequency} and the rest 13\% are due to the lack of required knowledge.
    Error analysis demonstrates that the model can be further improved by avoiding overfitting to the local context and incorporating explicit knowledge.
    Some erroneous cases are provided in Appendix~\ref{sec:appendix_error_case}.

    \section{Conclusion}\label{sec:conclusion}
    
    In this paper, we focus on grounding pronouns in dialogues to out-of-text objects.
    We propose to incorporate the topics of the dialogues to help the PCR model identify the out-of-text referents better.
    Experiments show that the proposed model outperforms previous models on both in-text and out-of-text PCR tasks.
    Detailed analysis is presented to show the strength and limitations of the proposed model.
    While this work is a first step to explore exophora resolution on one dataset, future work may explore exophora resolution in different domains such as AI chat-bots for home assistants.

    \section*{Acknowledgement}
    The authors of this paper were supported by the National Key Research and Development Program of China (No. 2018AAA0100701), a grant from the Guoqiang Institute, Tsinghua University, the NSFC Fund (U20B2053) from the NSFC of China, the RIF (R6020-19 and R6021-20) and the GRF (16211520) from RGC of Hong Kong, the MHKJFS (MHP/001/19) from ITC of Hong Kong, and the Tencent AI Lab Rhino-Bird Focused Research Program.

\bibliography{anthology,custom}

\begin{thebibliography}{31}
\expandafter\ifx\csname natexlab\endcsname\relax\def\natexlab#1{#1}\fi

\bibitem[{Akta{\c{s}} et~al.(2018)Akta{\c{s}}, Scheffler, and
  Stede}]{aktas-etal-2018-anaphora}
Berfin Akta{\c{s}}, Tatjana Scheffler, and Manfred Stede. 2018.
\newblock \href {https://doi.org/10.18653/v1/W18-0701} {Anaphora resolution for
  {T}witter conversations: An exploratory study}.
\newblock In \emph{Proceedings of the First Workshop on Computational Models of
  Reference, Anaphora and Coreference}, pages 1--10, New Orleans, Louisiana.

\bibitem[{Bahdanau et~al.(2015)Bahdanau, Cho, and
  Bengio}]{DBLP:journals/corr/BahdanauCB14}
Dzmitry Bahdanau, Kyunghyun Cho, and Yoshua Bengio. 2015.
\newblock \href {http://arxiv.org/abs/1409.0473} {Neural machine translation by
  jointly learning to align and translate}.
\newblock In \emph{Proceedings of ICLR 2015}.

\bibitem[{Bai et~al.(2021)Bai, Zhang, Song, and Xu}]{DBLP:conf/eacl/BaiZSX21}
Jiaxin Bai, Hongming Zhang, Yangqiu Song, and Kun Xu. 2021.
\newblock \href {https://aclanthology.org/2021.eacl-main.43/} {Joint
  coreference resolution and character linking for multiparty conversation}.
\newblock In \emph{EACL}, pages 539--548. Association for Computational
  Linguistics.

\bibitem[{Bird(2006)}]{DBLP:conf/acl/Bird06}
Steven Bird. 2006.
\newblock \href {https://aclanthology.org/P06-4018/} {{NLTK:} the natural
  language toolkit}.
\newblock In \emph{Proceedings of ACL 2006}.

\bibitem[{Blei et~al.(2001)Blei, Ng, and Jordan}]{DBLP:conf/nips/BleiNJ01}
David~M. Blei, Andrew~Y. Ng, and Michael~I. Jordan. 2001.
\newblock \href
  {https://proceedings.neurips.cc/paper/2001/hash/296472c9542ad4d4788d543508116cbc-Abstract.html}
  {Latent dirichlet allocation}.
\newblock In \emph{Proceedings of NIPS 2001}, pages 601--608.

\bibitem[{Brown and Yule(1983)}]{brown1983discourse}
Gillian Brown and George Yule. 1983.
\newblock \emph{Discourse analysis}.
\newblock Cambridge university press.

\bibitem[{Das et~al.(2017)Das, Kottur, Gupta, Singh, Yadav, Moura, Parikh, and
  Batra}]{DBLP:conf/cvpr/DasKGSYMPB17}
Abhishek Das, Satwik Kottur, Khushi Gupta, Avi Singh, Deshraj Yadav, Jos{\'{e}}
  M.~F. Moura, Devi Parikh, and Dhruv Batra. 2017.
\newblock \href {https://doi.org/10.1109/CVPR.2017.121} {Visual dialog}.
\newblock In \emph{Proceedings of CVPR 2017}, pages 1080--1089.

\bibitem[{Dasigi et~al.(2019)Dasigi, Liu, Marasovic, Smith, and
  Gardner}]{DBLP:conf/emnlp/DasigiLMSG19}
Pradeep Dasigi, Nelson~F. Liu, Ana Marasovic, Noah~A. Smith, and Matt Gardner.
  2019.
\newblock \href {https://doi.org/10.18653/v1/D19-1606} {Quoref: {A} reading
  comprehension dataset with questions requiring coreferential reasoning}.
\newblock In \emph{Proceedings of EMNLP-IJCNLP}, pages 5924--5931. Association
  for Computational Linguistics.

\bibitem[{Ehrlich(1981)}]{DBLP:conf/acl/Ehrlich81}
Kate Ehrlich. 1981.
\newblock \href {https://aclanthology.org/P81-1019/} {Search and inference
  strategies in pronoun resolution: an experimental study}.
\newblock In \emph{Proceedings of ACL 1981}, pages 89--93.

\bibitem[{Guillou(2012)}]{DBLP:conf/eacl/Guillou12}
Liane Guillou. 2012.
\newblock \href {https://aclanthology.org/E12-3001/} {Improving pronoun
  translation for statistical machine translation}.
\newblock In \emph{Proceedings of EACL 2012}, pages 1--10.

\bibitem[{Halliday and Hasan(1976)}]{halliday1976cohesion}
M.A.K. Halliday and R.~Hasan. 1976.
\newblock Cohesion in english.
\newblock \emph{Longman}, pages 18--33.

\bibitem[{Hangyo et~al.(2013)Hangyo, Kawahara, and
  Kurohashi}]{DBLP:conf/emnlp/HangyoKK13}
Masatsugu Hangyo, Daisuke Kawahara, and Sadao Kurohashi. 2013.
\newblock \href {https://aclanthology.org/D13-1095/} {Japanese zero reference
  resolution considering exophora and author/reader mentions}.
\newblock In \emph{EMNLP}, pages 924--934. {ACL}.

\bibitem[{Hankamer and Sag(1976)}]{hankamer1976deep}
Jorge Hankamer and Ivan Sag. 1976.
\newblock Deep and surface anaphora.
\newblock \emph{Linguistic inquiry}, 7(3):391--428.

\bibitem[{Hobbs(1978)}]{hobbs1978resolving}
Jerry~R Hobbs. 1978.
\newblock Resolving pronoun references.
\newblock \emph{Lingua}, 44(4):311--338.

\bibitem[{Joshi et~al.(2020)Joshi, Chen, Liu, Weld, Zettlemoyer, and
  Levy}]{DBLP:journals/tacl/JoshiCLWZL20}
Mandar Joshi, Danqi Chen, Yinhan Liu, Daniel~S. Weld, Luke Zettlemoyer, and
  Omer Levy. 2020.
\newblock \href {https://transacl.org/ojs/index.php/tacl/article/view/1853}
  {Spanbert: Improving pre-training by representing and predicting spans}.
\newblock \emph{TACL}, 8:64--77.

\bibitem[{Joshi et~al.(2019)Joshi, Levy, Zettlemoyer, and
  Weld}]{DBLP:conf/emnlp/JoshiLZW19}
Mandar Joshi, Omer Levy, Luke Zettlemoyer, and Daniel~S. Weld. 2019.
\newblock \href {https://doi.org/10.18653/v1/D19-1588} {{BERT} for coreference
  resolution: Baselines and analysis}.
\newblock In \emph{Proceedings of EMNLP-IJCNLP 2019}, pages 5802--5807.

\bibitem[{Kottur et~al.(2018)Kottur, Moura, Parikh, Batra, and
  Rohrbach}]{DBLP:conf/eccv/KotturMPBR18}
Satwik Kottur, Jos{\'{e}} M.~F. Moura, Devi Parikh, Dhruv Batra, and Marcus
  Rohrbach. 2018.
\newblock \href {https://doi.org/10.1007/978-3-030-01267-0\_10} {Visual
  coreference resolution in visual dialog using neural module networks}.
\newblock In \emph{Proceedings of ECCV 2018}, pages 160--178.

\bibitem[{Lee et~al.(2018)Lee, He, and Zettlemoyer}]{lee-etal-2018-higher}
Kenton Lee, Luheng He, and Luke Zettlemoyer. 2018.
\newblock \href {https://doi.org/10.18653/v1/n18-2108} {Higher-order
  coreference resolution with coarse-to-fine inference}.
\newblock In \emph{Proceedings of NAACL-HLT 2018}, pages 687--692.

\bibitem[{Miller(1995)}]{DBLP:journals/cacm/Miller95}
George~A. Miller. 1995.
\newblock \href {http://doi.acm.org/10.1145/219717.219748} {Wordnet: {A}
  lexical database for english}.
\newblock \emph{Commun. {ACM}}, 38(11):39--41.

\bibitem[{Mitkov(1998)}]{mitkov1998robust}
Ruslan Mitkov. 1998.
\newblock \href {https://aclanthology.org/P98-2143/} {Robust pronoun resolution
  with limited knowledge}.
\newblock In \emph{Proceedings of ACL 1998}, pages 869--875.

\bibitem[{Ng(2005)}]{DBLP:conf/aaai/Ng05}
Vincent Ng. 2005.
\newblock \href {http://www.aaai.org/Library/AAAI/2005/aaai05-171.php}
  {Supervised ranking for pronoun resolution: Some recent improvements}.
\newblock In \emph{Proceedings of AAAI 2005}, pages 1081--1086.

\bibitem[{NIST(2003)}]{nist2003ace}
US~NIST. 2003.
\newblock The ace 2003 evaluation plan.
\newblock \emph{US National Institute for Standards and Technology (NIST)},
  pages 2003--08.

\bibitem[{Niu et~al.(2019)Niu, Zhang, Zhang, Zhang, Lu, and
  Wen}]{DBLP:conf/cvpr/NiuZZZLW19}
Yulei Niu, Hanwang Zhang, Manli Zhang, Jianhong Zhang, Zhiwu Lu, and Ji{-}Rong
  Wen. 2019.
\newblock \href
  {http://openaccess.thecvf.com/content\_CVPR\_2019/html/Niu\_Recursive\_Visual\_Attention\_in\_Visual\_Dialog\_CVPR\_2019\_paper.html}
  {Recursive visual attention in visual dialog}.
\newblock In \emph{Proceedings of CVPR 2019}, pages 6679--6688.

\bibitem[{Pradhan et~al.(2012)Pradhan, Moschitti, Xue, Uryupina, and
  Zhang}]{pradhan2012conll}
Sameer Pradhan, Alessandro Moschitti, Nianwen Xue, Olga Uryupina, and Yuchen
  Zhang. 2012.
\newblock \href {https://aclanthology.org/W12-4501/} {Conll-2012 shared task:
  Modeling multilingual unrestricted coreference in ontonotes}.
\newblock In \emph{Proceedings of EMNLP-CoNLL 2012}, pages 1--40.

\bibitem[{Steinberger et~al.(2007)Steinberger, Poesio, Kabadjov, and
  Jezek}]{DBLP:journals/ipm/SteinbergerPKJ07}
Josef Steinberger, Massimo Poesio, Mijail~A. Kabadjov, and Karel Jezek. 2007.
\newblock \href {https://doi.org/10.1016/j.ipm.2007.01.010} {Two uses of
  anaphora resolution in summarization}.
\newblock \emph{Inf. Process. Manag.}, 43(6):1663--1680.

\bibitem[{Strube and M{\"{u}}ller(2003)}]{DBLP:conf/acl/StrubeM03}
Michael Strube and Christoph M{\"{u}}ller. 2003.
\newblock \href {https://aclanthology.org/P03-1022/} {A machine learning
  approach to pronoun resolution in spoken dialogue}.
\newblock In \emph{Proceedings of ACL 2003}, pages 168--175.

\bibitem[{Wong et~al.(2020)Wong, Maruf, and Haffari}]{DBLP:conf/acl/WongMH20}
KayYen Wong, Sameen Maruf, and Gholamreza Haffari. 2020.
\newblock \href {https://doi.org/10.18653/v1/2020.acl-main.530} {Contextual
  neural machine translation improves translation of cataphoric pronouns}.
\newblock In \emph{Proceedings of ACL 2020}, pages 5971--5978.

\bibitem[{Wu et~al.(2020)Wu, Wang, Yuan, Wu, and Li}]{DBLP:conf/acl/WuWYWL20}
Wei Wu, Fei Wang, Arianna Yuan, Fei Wu, and Jiwei Li. 2020.
\newblock \href {https://doi.org/10.18653/v1/2020.acl-main.622} {Corefqa:
  Coreference resolution as query-based span prediction}.
\newblock In \emph{Proceedings of ACL 2020}, pages 6953--6963.

\bibitem[{Yu et~al.(2019)Yu, Zhang, Song, Song, and
  Zhang}]{DBLP:conf/emnlp/YuZSSZ19}
Xintong Yu, Hongming Zhang, Yangqiu Song, Yan Song, and Changshui Zhang. 2019.
\newblock \href {https://doi.org/10.18653/v1/D19-1516} {What you see is what
  you get: Visual pronoun coreference resolution in dialogues}.
\newblock In \emph{Proceedings of EMNLP-IJCNLP 2019}, pages 5122--5131.

\bibitem[{Yule(1979)}]{yule1979pragmatically}
George Yule. 1979.
\newblock Pragmatically controlled anaphora.
\newblock \emph{Lingua}, 49(2-3):127--135.

\bibitem[{Zhang et~al.(2019)Zhang, Song, and Song}]{DBLP:conf/naacl/ZhangSS19}
Hongming Zhang, Yan Song, and Yangqiu Song. 2019.
\newblock \href {https://doi.org/10.18653/v1/n19-1093} {Incorporating context
  and external knowledge for pronoun coreference resolution}.
\newblock In \emph{Proceedings of NAACL-HLT 2019}, pages 872--881.

\end{thebibliography}
\bibliographystyle{acl_natbib}

\clearpage

\appendix

\section{Task Definition Compared to Prior Works}
\label{sec:appendix}
\begin{figure*}
    \centering
    \subfigure[]{\label{fig:VD_example_setting}
        \includegraphics[width=0.55\textwidth]{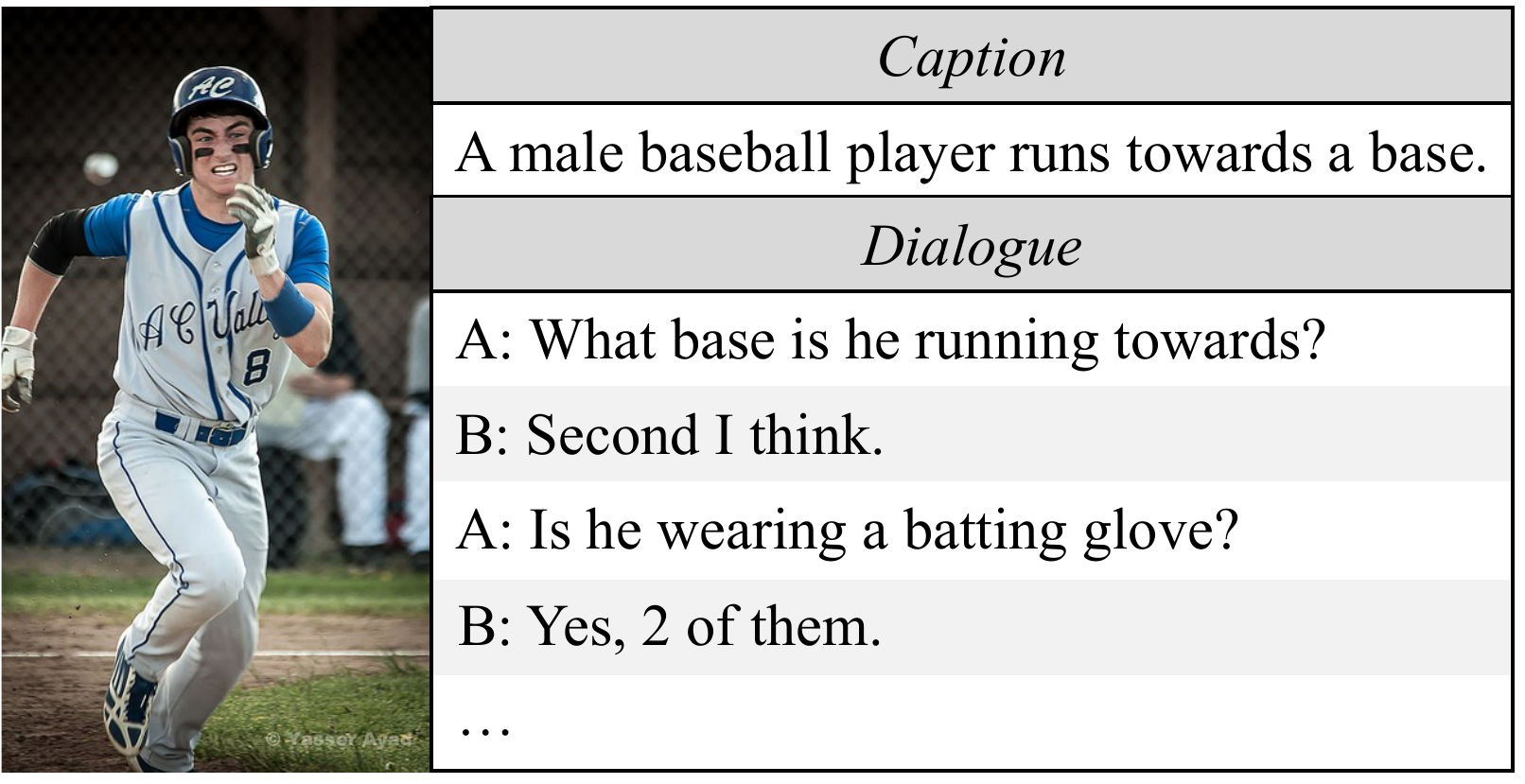}
    }
    \subfigure[]{\label{fig:VPCR_example_setting}
        \includegraphics[width=0.85\textwidth]{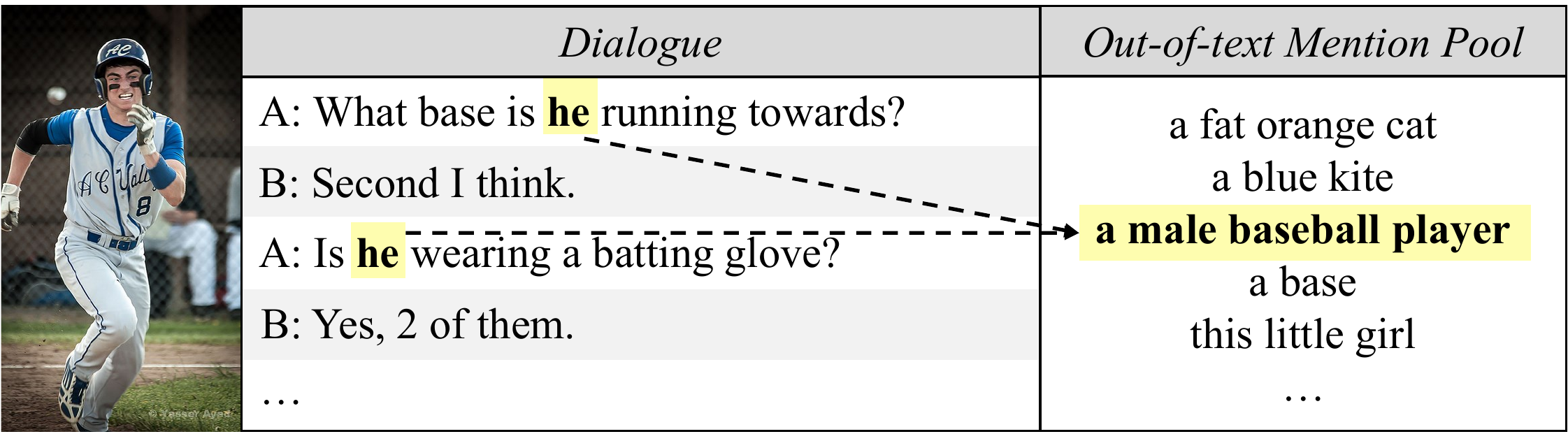}
    }
    \subfigure[]{\label{fig:Exo_example_setting}
        \includegraphics[width=0.7\textwidth]{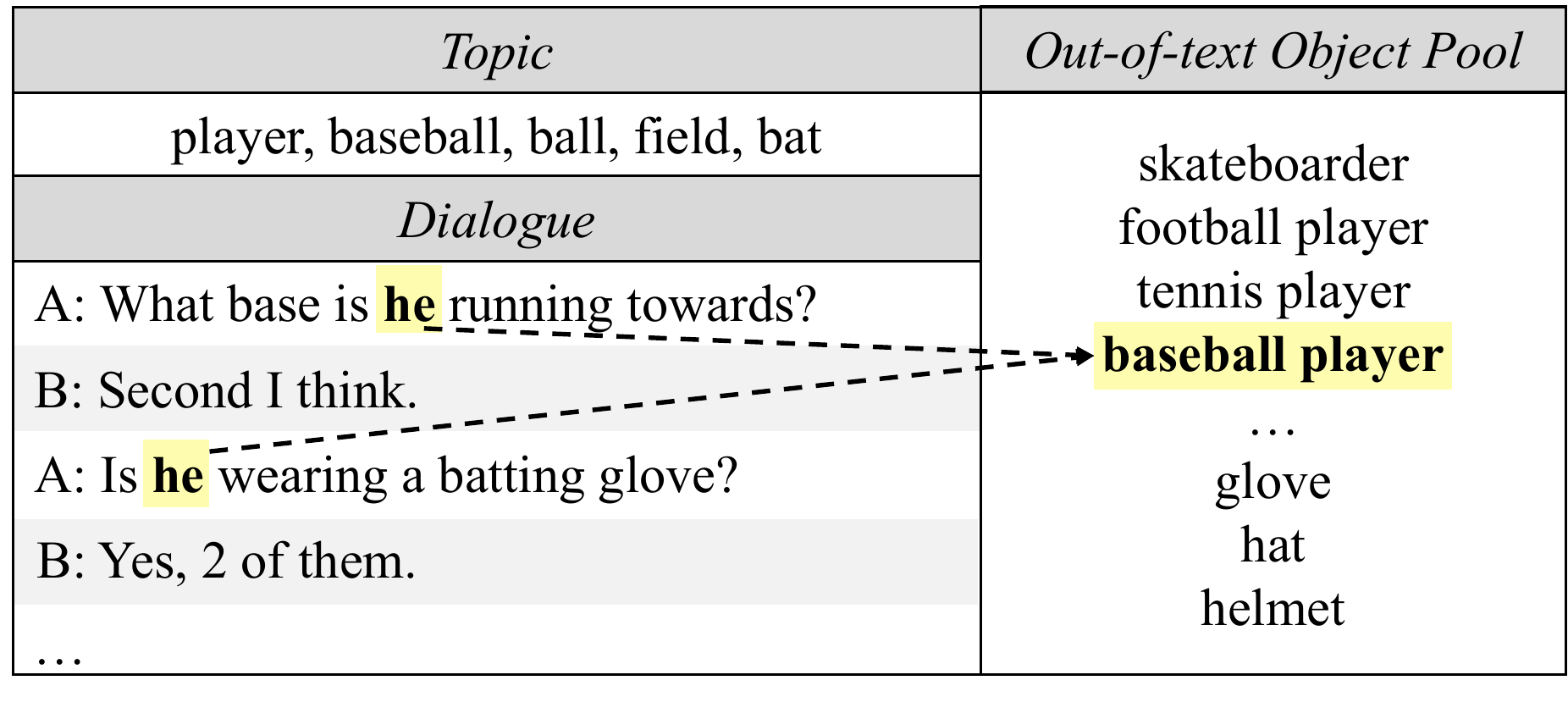}
    }
    \caption{Examples of different settings in (a) Visual Dialog, (b) VisPro, and (c) ours.
    }
    \label{fig:setting}
\end{figure*}

    Our experiments are based on the dataset VisPro~\cite{DBLP:conf/emnlp/YuZSSZ19}, which provides annotation of referents for pronouns in dialogues of the Visual Dialog dataset~\cite{DBLP:conf/cvpr/DasKGSYMPB17}. 
    Figure~\ref{fig:setting} illustrates the different settings of our work compared to prior works.
    
    In the original setting of Visual Dialog dataset (Figure~\ref{fig:VD_example_setting}), each dialogue happens between two people chatting about an image, and each image is accompanied by a descriptive caption. Speaker A only has access to the caption and attempts to imagine the image by asking questions, while speaker B can access both the image and the caption and answers the questions. Thus the pronouns in the dialogues refer to either mention in the dialogue text or noun phrase in captions.

    In the setting of VisPro (Figure~\ref{fig:VPCR_example_setting}), to simulate the scenario where people use pronouns to directly refer to objects in the environment, the captions are separated from the dialogues. As the captions are descriptions of images, the mentions in captions must correspond to some objects in the images. Thus, when captions are no longer available, the pronouns that refer to noun phrases in captions can be deemed as referring to objects in the images.

    Although VisPro first proposed the scenario where pronouns refer to out-of-text objects, it focused on visual-related cases and did not associate such cases with the general definition of exophora. Furthermore, in the definition of the visual pronoun coreference resolution task that \citet{DBLP:conf/emnlp/YuZSSZ19} proposed, the candidates of the out-of-text objects are 30 noun phrases randomly selected from captions. This small set of objects contains noun phrases from the corresponding caption as well as captions of other images to provide negative samples. However, such a setting is not so practical. For one thing, the out-of-text candidates are not fixed among different dialogues and the choices for negative samples are random, which makes it hard to compare between multiple dialogues. For another, such noun phrases are hard to obtain in practical cases where no caption for the environment is available, so the model trained under this task cannot be applied to dialogues outside the dataset.

    Based on the annotation of VisPro, we design a more practical experiment setting (Figure~\ref{fig:Exo_example_setting}). First, we assume that the visual background of dialogues is not always available, and thus aim to resolve exophoric pronouns based on only the dialogue text. Second, since exophoric pronouns might refer to any object in daily life, we collect all the common objects in VisPro to form a candidate pool of 384 object categories. 
    Since the candidate pool is fixed for all dialogues, we can reasonably compare the performance between different dialogues and models. The model trained under our setting can thus be applied to real-life dialogues.

\section{Case Study for Out-of-Text PCR}
\label{sec:appendix_case}

\begin{figure*}[t]
        \centering
        \subfigure[]{\label{fig:acase1}
            \includegraphics[width=0.48\textwidth]{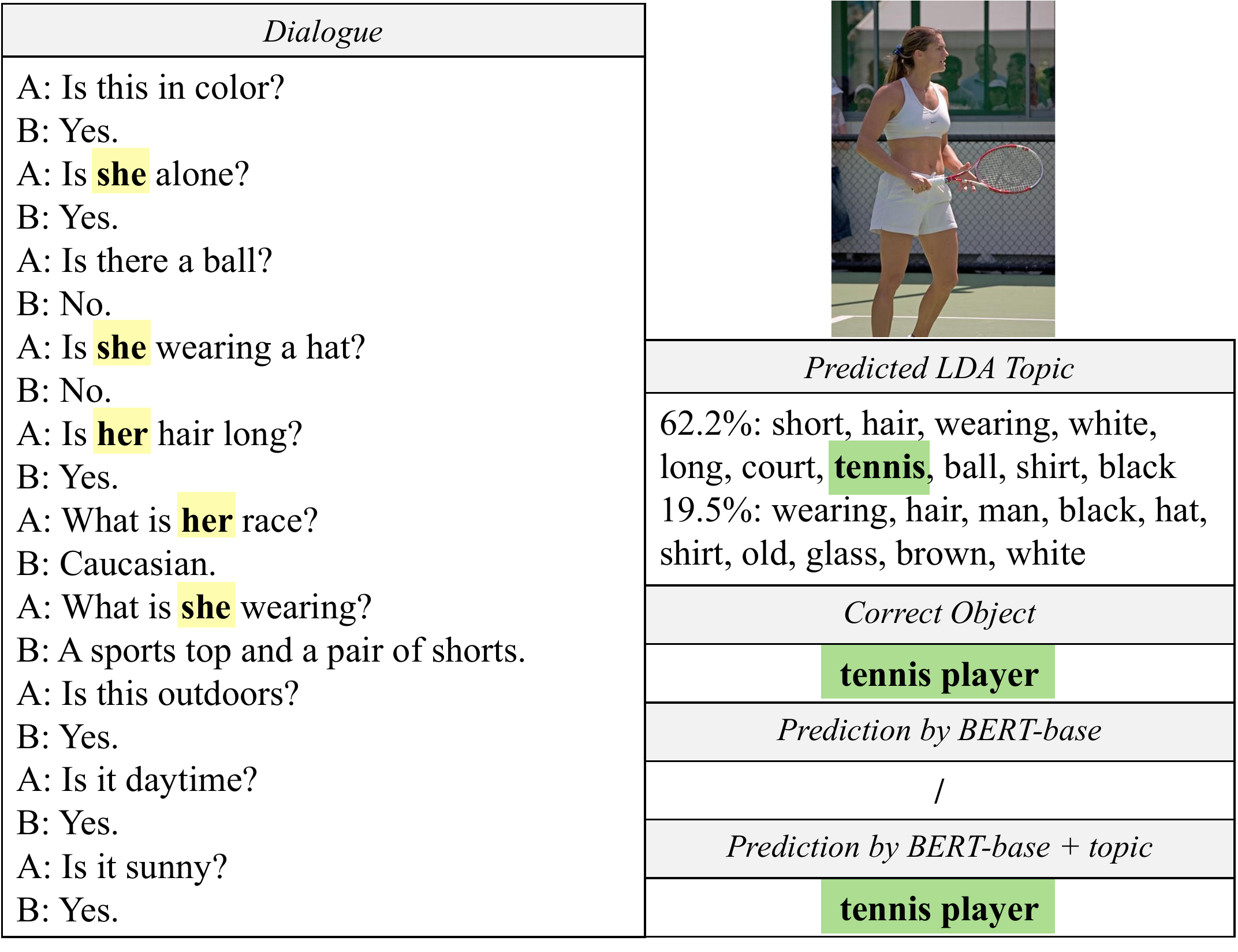}
        }
        \subfigure[]{\label{fig:acase2}
            \includegraphics[width=0.48\textwidth]{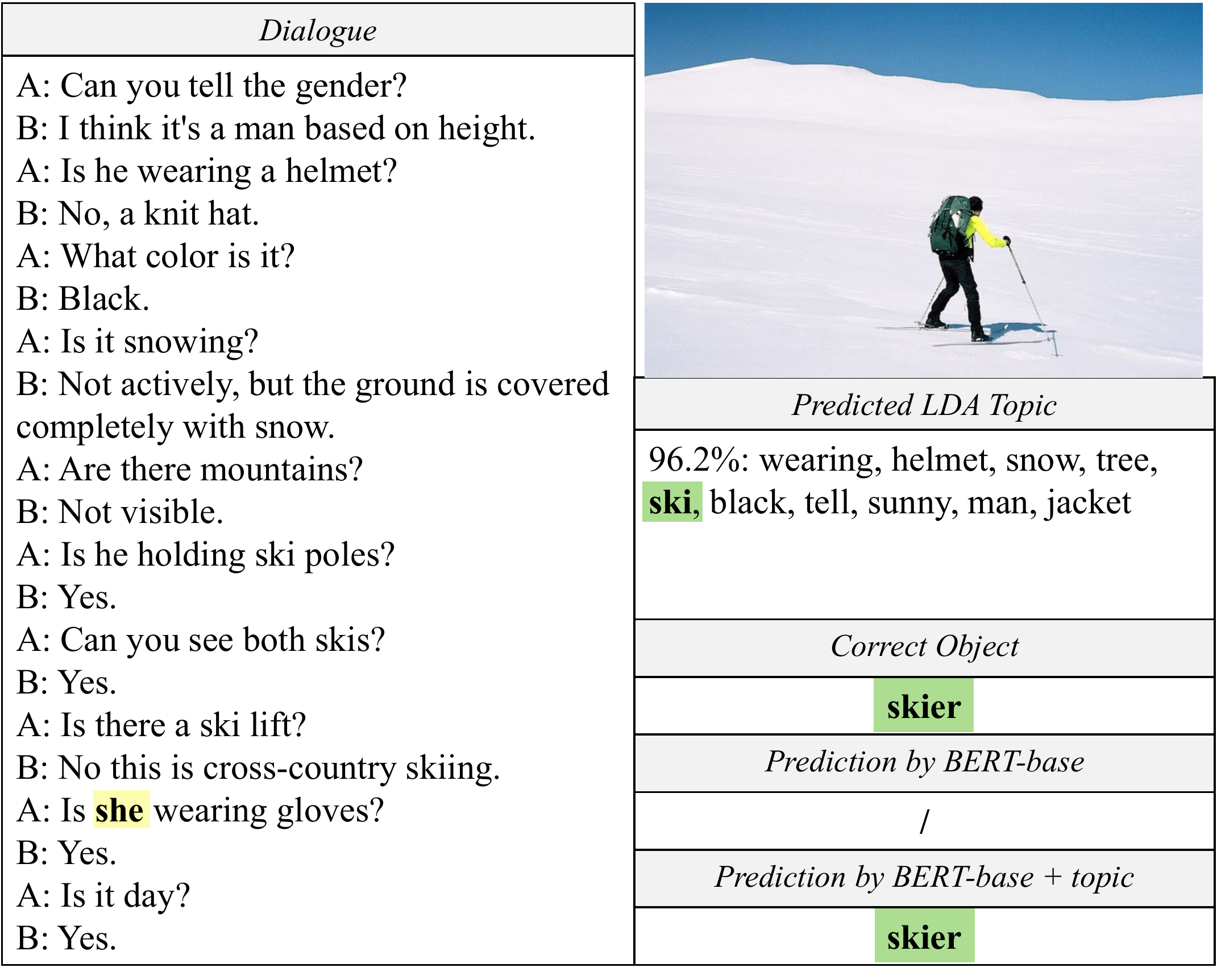}
        }
        \subfigure[]{\label{fig:acase3}
            \includegraphics[width=0.48\textwidth]{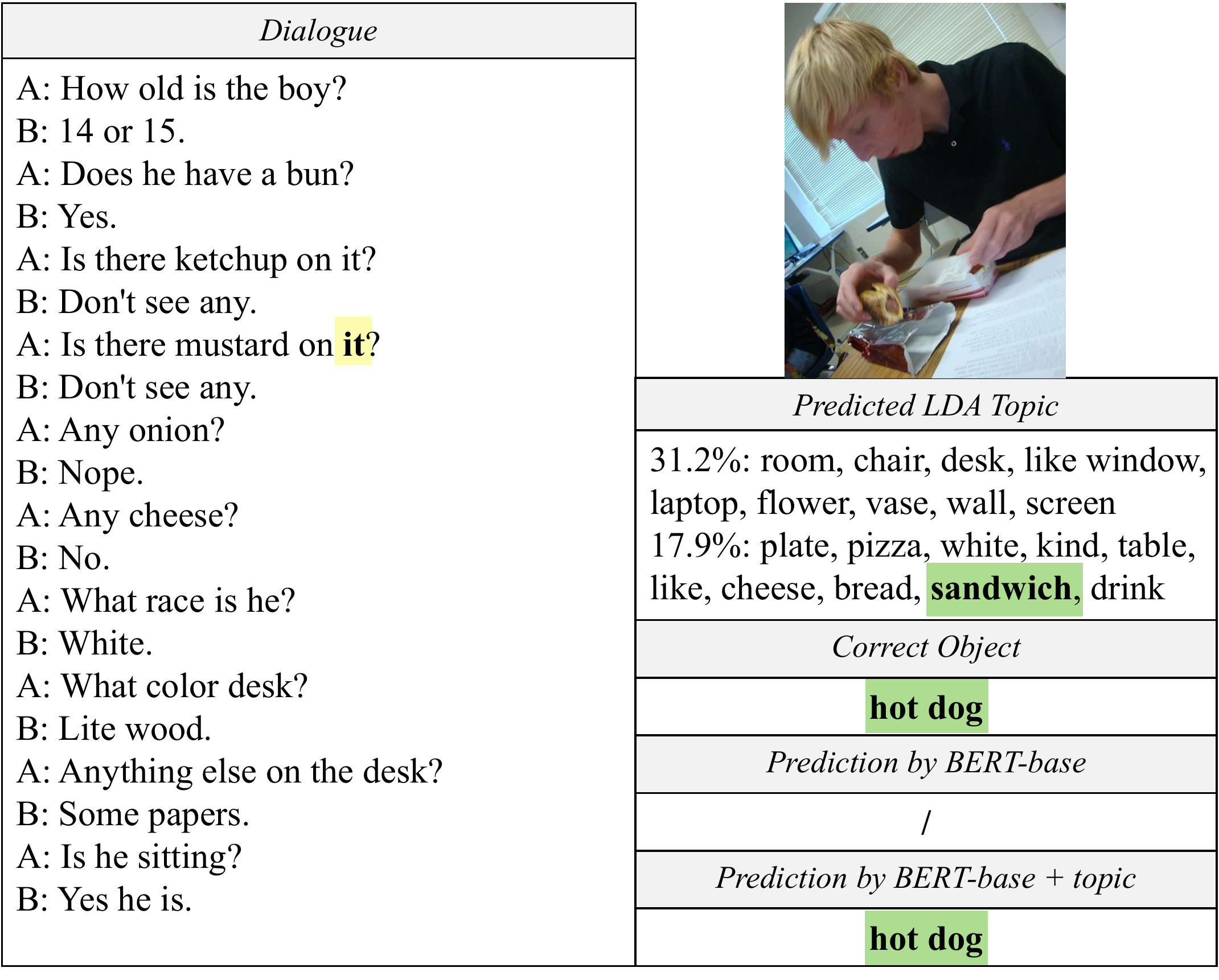}
        }
        \subfigure[]{\label{fig:acase4}
            \includegraphics[width=0.48\textwidth]{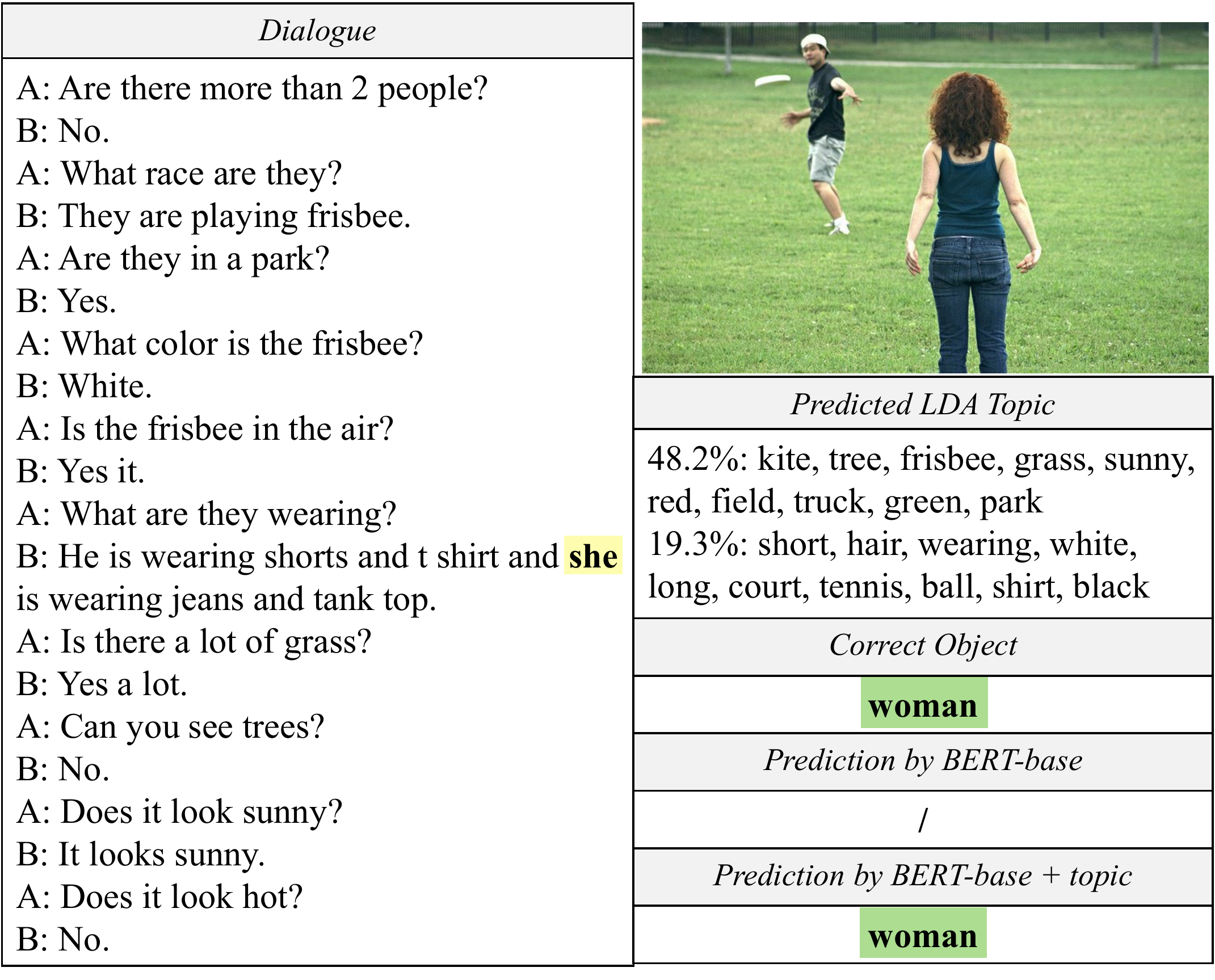}
        }
        \subfigure[]{\label{fig:acase5}
            \includegraphics[width=0.48\textwidth]{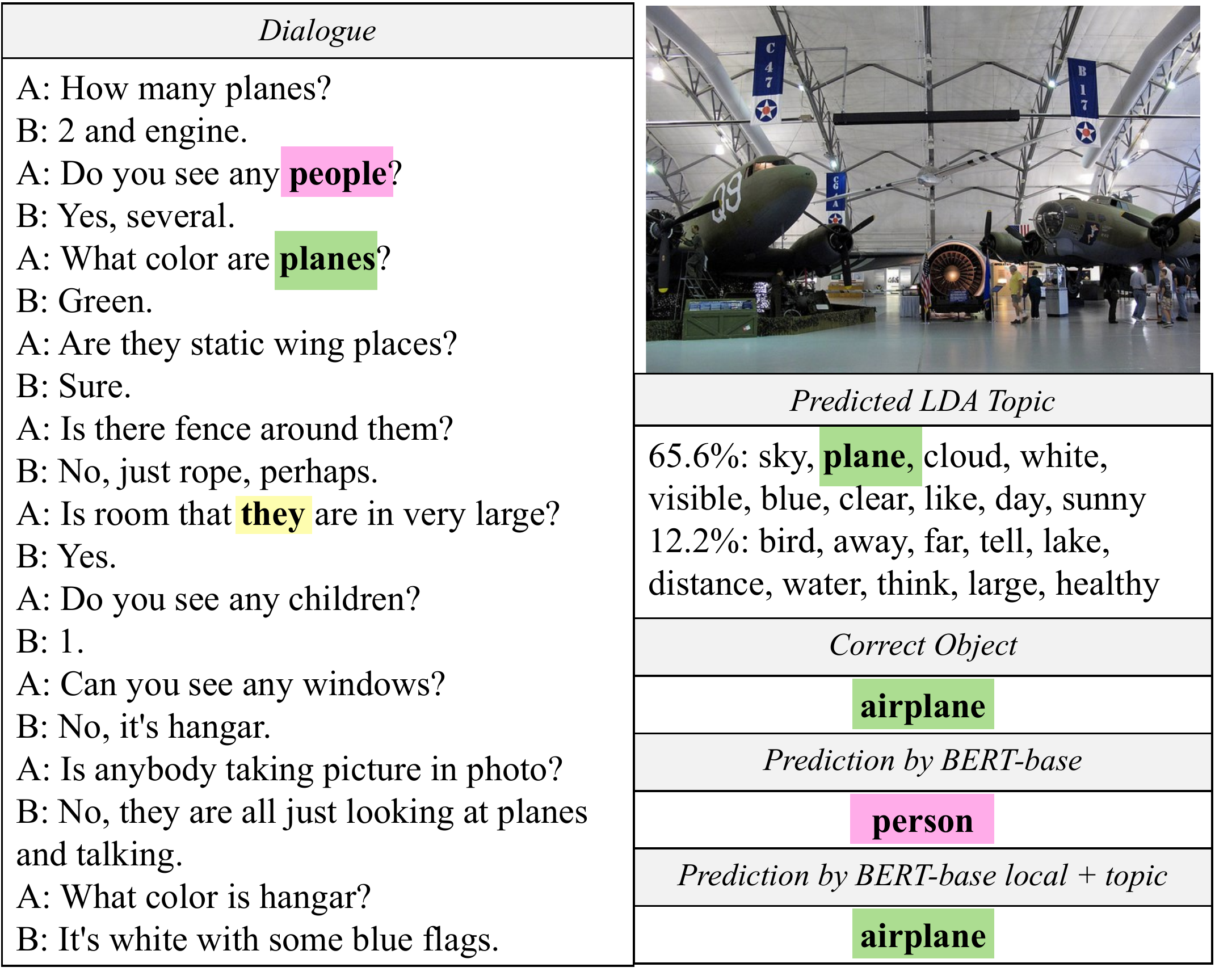}
        }
        \subfigure[]{\label{fig:acase6}
            \includegraphics[width=0.48\textwidth]{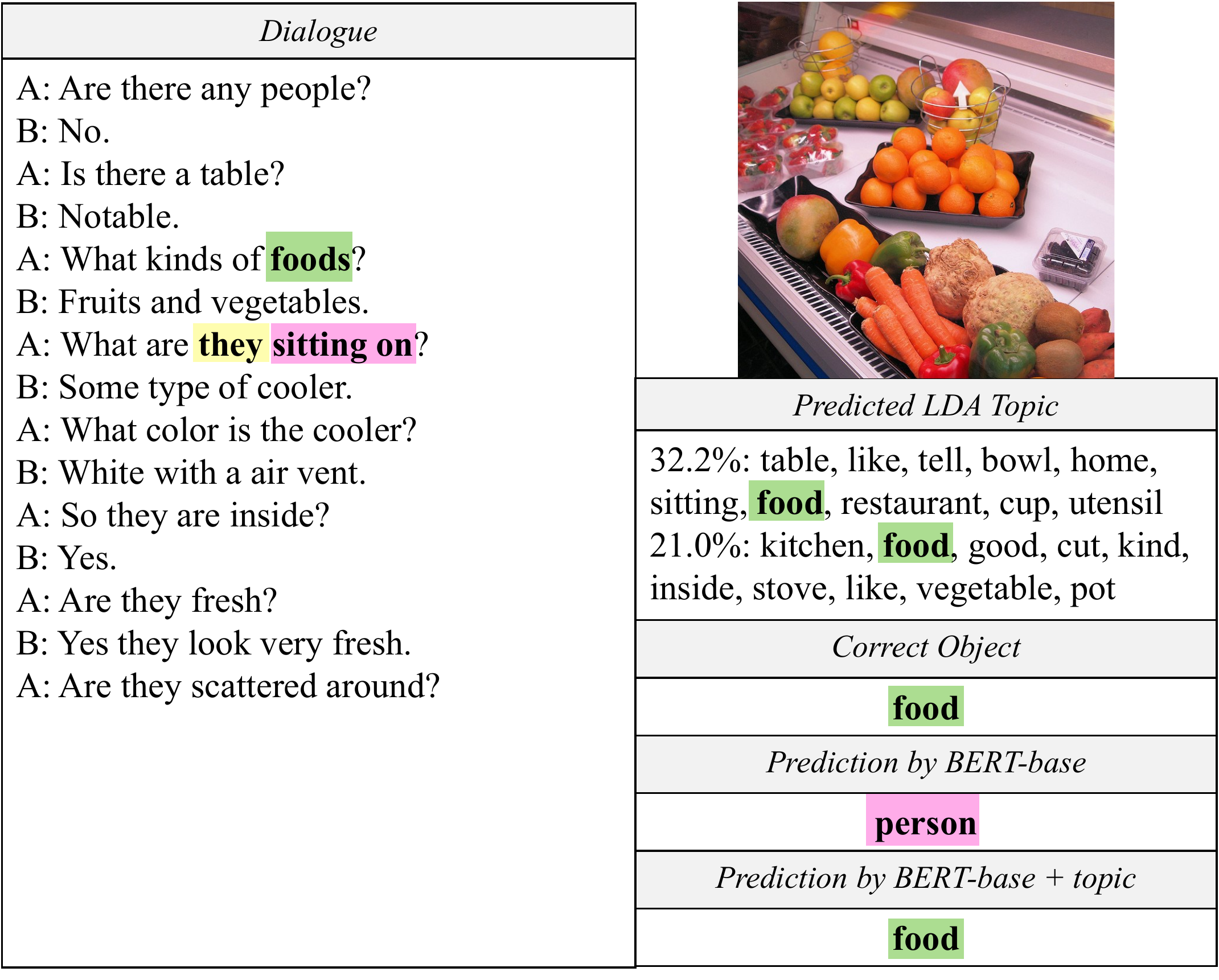}
        }
        \caption{Case study for (a)-(d) ``Not Discussed'' and (e)(f) ``Discussed'' out-of-text PCR. 
        \hlc[yellow]{Target pronouns}, \hlc[dark-green]{correct out-of-text objects} with their hints, and \hlc[purple-pink]{false prediction} with distracting words are marked in different colors. 
        Note that the images are not provided to the models.
        }
        \label{fig:acase}
    \end{figure*}

    We randomly select some cases from the test split of VisPro and present them in Figure~\ref{fig:acase}. Cases (a)-(d) are ``Not Discussed'' pronouns which only have out-of-text referents, and cases (e)(f) are ``Discussed'' pronouns which have both out-of-text and in-text referents. For the ``Discussed'' pronouns in (e)(f), even though the referred objects are mentioned in the text, the BERT-base model still overfits to distracting words and gives the false prediction ``person.'' On the contrary, our model leverages the topic information and predicts the correct objects.
    
\section{Erroneous Case Study for Out-of-Text PCR}
\label{sec:appendix_error_case}
\begin{figure*}[t]
        \centering
        \subfigure[]{\label{fig:ecase1}
            \includegraphics[width=0.48\textwidth]{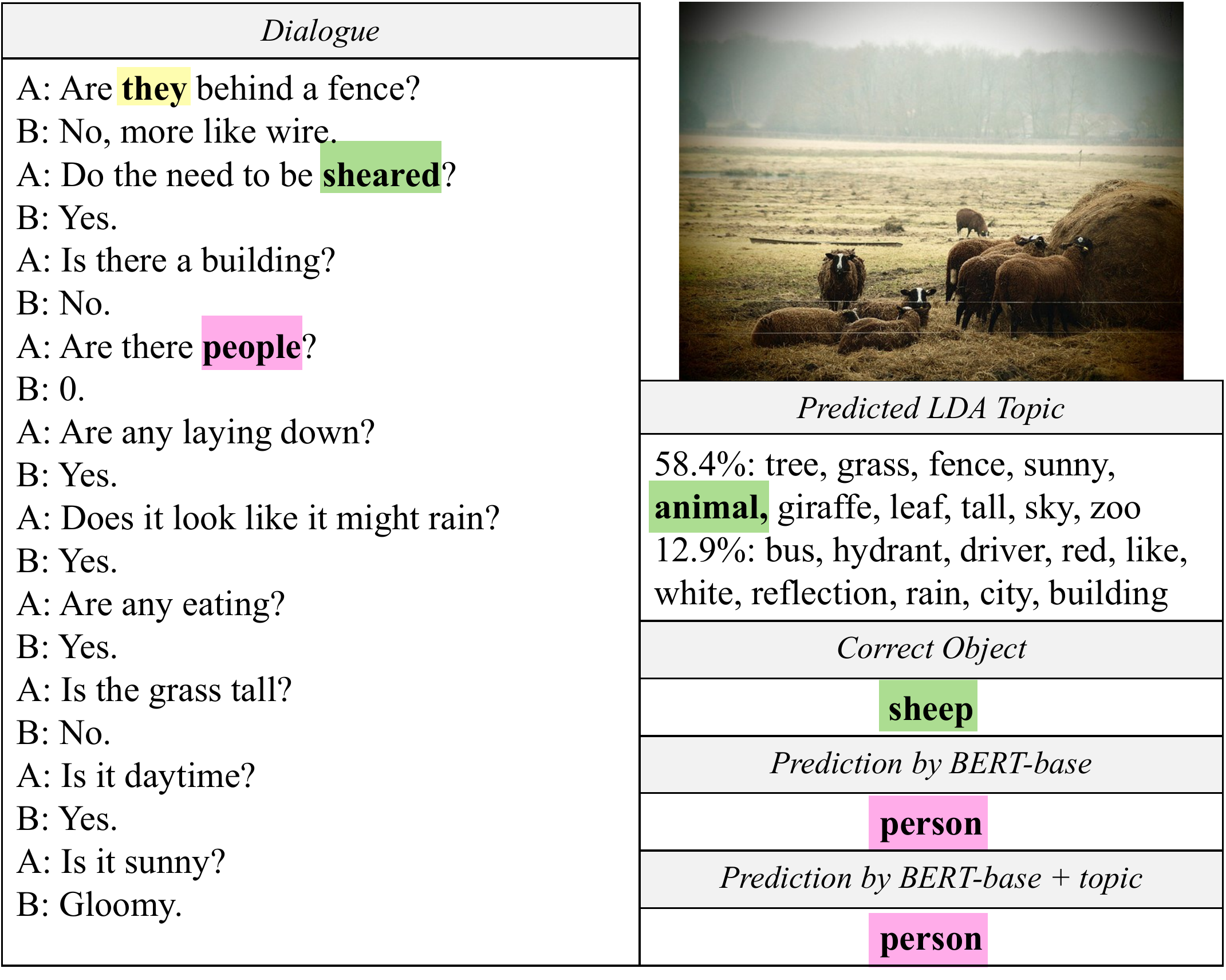}
        }
        \subfigure[]{\label{fig:ecase2}
            \includegraphics[width=0.48\textwidth]{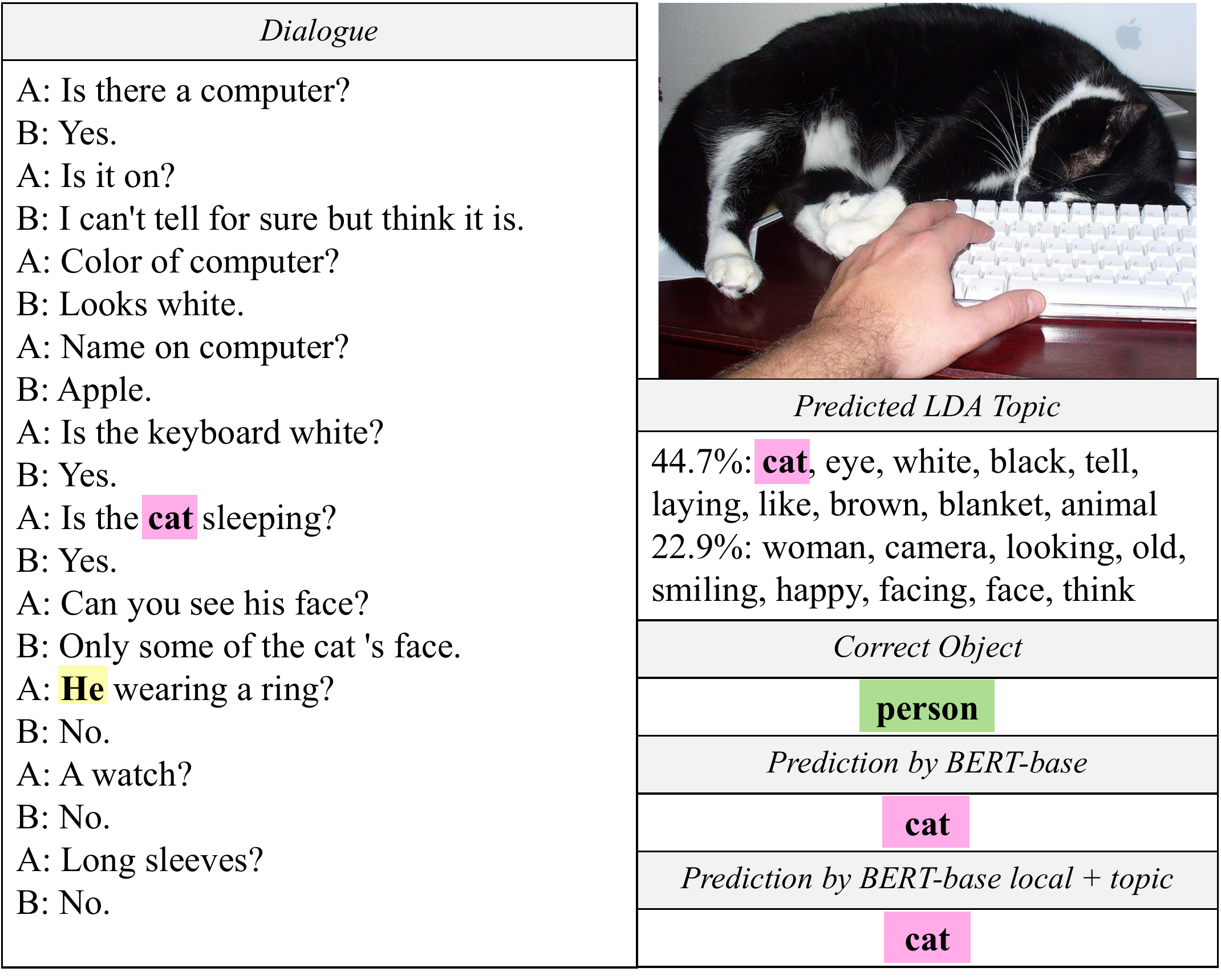}
        }
        \caption{Erroneous case study for ``Not Discussed'' out-of-text PCR. 
        \hlc[yellow]{Target pronouns}, \hlc[dark-green]{correct out-of-text objects} with their hints, and \hlc[purple-pink]{false prediction} with distracting words are marked in different colors. 
        Note that the images are not provided to the models.
        }
        \label{fig:ecase}
    \end{figure*}

    Figure~\ref{fig:ecase} presents some typical erroneous cases.
    In Figure~\ref{fig:ecase1}, the model predicts that ``they'' refers to ``person'' instead of ``sheep,'' which hits three error types.
    First, the topic model correctly infers that the dialogue is about some animals on grass but the coreference model ignores this information.
    Second, based on the word ``sheared'' and knowledge that sheep need to be sheared, humans can infer that the pronoun refers to ``sheep.''
    However, the model fails to learn such knowledge from the pre-training of the language model.
    Last, the prediction of ``person'' indicates that it overfits to the word ``people'' in dialogue text even though it says that there are 0 people.
    Figure~\ref{fig:ecase2} shows another case where the model fails to recall the knowledge that only a person could wear a ring or a watch and thus fail to infer that ``he'' refers to a person.
    
\end{document}